\pdfoutput=1

\documentclass[sigconf]{acmart}

\copyrightyear{2026}
\acmYear{2026}
\setcopyright{cc}
\setcctype{by-nc-nd}
\acmConference[MM '26]{Proceedings of the 34th ACM International Conference on Multimedia}{November 10--14, 2026}{Rio de Janeiro, Brazil}
\acmBooktitle{Proceedings of the 34th ACM International Conference on Multimedia (MM '26), November 10--14, 2026, Rio de Janeiro, Brazil}
\acmDOI{10.1145/3767308.3836198}
\acmISBN{979-8-4007-2213-4/2026/11}

\usepackage{float}
\usepackage{subcaption}
\usepackage{multirow}
\usepackage{pifont}
\newcommand{\xmark}{\ding{55}} 
\usepackage{array}
\newcommand{\yes}{\ding{51}}
\newcommand{\no}{\ding{55}}
\AtBeginDocument{
  \setlength{\abovedisplayskip}{5pt plus 1pt minus 1pt}
  \setlength{\belowdisplayskip}{5pt plus 1pt minus 1pt}
  \setlength{\abovedisplayshortskip}{3pt plus 1pt minus 1pt}
  \setlength{\belowdisplayshortskip}{3pt plus 1pt minus 1pt}

  \setlength{\textfloatsep}{5pt plus 1pt minus 1pt}
  \setlength{\floatsep}{5pt plus 1pt minus 1pt}
  \setlength{\intextsep}{5pt plus 1pt minus 1pt}

  \setlength{\abovecaptionskip}{5pt}
  \setlength{\belowcaptionskip}{2pt}

}
\begin{document}

\title{GeoMAR: Unleashing Geometrically Aligned Features for Masked Autoregressive Blind Face Restoration}

\author{Lu Gan}
\authornote{Both authors contributed equally to this research.}
\affiliation{%
  \institution{School of Artificial Intelligence, \\Sun Yat-sen University}
  \city{Zhuhai}
  \country{China}
}
\email{ganlu33@mail2.sysu.edu.cn}
\author{Hanyu Yan}
\authornotemark[1]
\affiliation{%
  \institution{School of Artificial Intelligence, \\Sun Yat-sen University}
  \city{Zhuhai}
  \country{China}}
\email{yanhy27@mail2.sysu.edu.cn}

\author{Chaofeng Chen}
  \affiliation{%
  \institution{Institute for Math and AI, Wuhan\\School of Artificial Intelligence, Wuhan University}
  \city{Wuhan}
  \country{China}
}
\email{chaofengchen@whu.edu.cn}

\author{Junqi Hu}
\affiliation{%
 \institution{School of Artificial Intelligence, \\Sun Yat-sen University}
 \city{Zhuhai}
 \country{China}}
\email{hujq28@mail2.sysu.edu.cn}

\author{Dan Zeng}
\authornote{Corresponding author: zengd8@mail.sysu.edu.cn.}
\affiliation{%
  \institution{School of Artificial Intelligence, \\Sun Yat-sen University}
  \city{Zhuhai}
  \country{China}}
\affiliation{%
  \institution{The Technology Innovation Center for Collaborative Applications of Natural Resources Data in GBA, MNR}
  \city{Guangzhou}
  \country{China}}
\email{zengd8@mail.sysu.edu.cn}


\begin{abstract}
      Codebook-based blind face restoration (BFR) often suffers from ambiguous conditioning features and a fragile prediction mechanism under severe degradation. To address these challenges, we propose GeoMAR, a framework designed to unleash geometrically aligned features with masked autoregressive (MAR) refinement for robust face restoration. For feature conditioning, we introduce a dual-input extraction pipeline to extract component-based geometric descriptions with explicit, spatially faithful anchors. These textual priors are integrated with low-quality (LQ) features via an Aligned Geometric Priors Injector, which employs a KV-Q exchange strategy to generate geometrically aligned features. For prediction mechanism, we reformulate the one-step mapping into a multi-step MAR process. This coarse-to-fine generation progressively refines complex facial regions based on increasingly reliable context. Experiments on one synthetic and three real-world benchmarks demonstrate that GeoMAR achieves highly competitive perceptual quality and coherent visual structures compared with existing methods. The code is available at \textcolor{purple}{\url{https://github.com/BRL-SYSU/GeoMAR.git}}.
\end{abstract}



\begin{teaserfigure}
 \includegraphics[width=\textwidth]{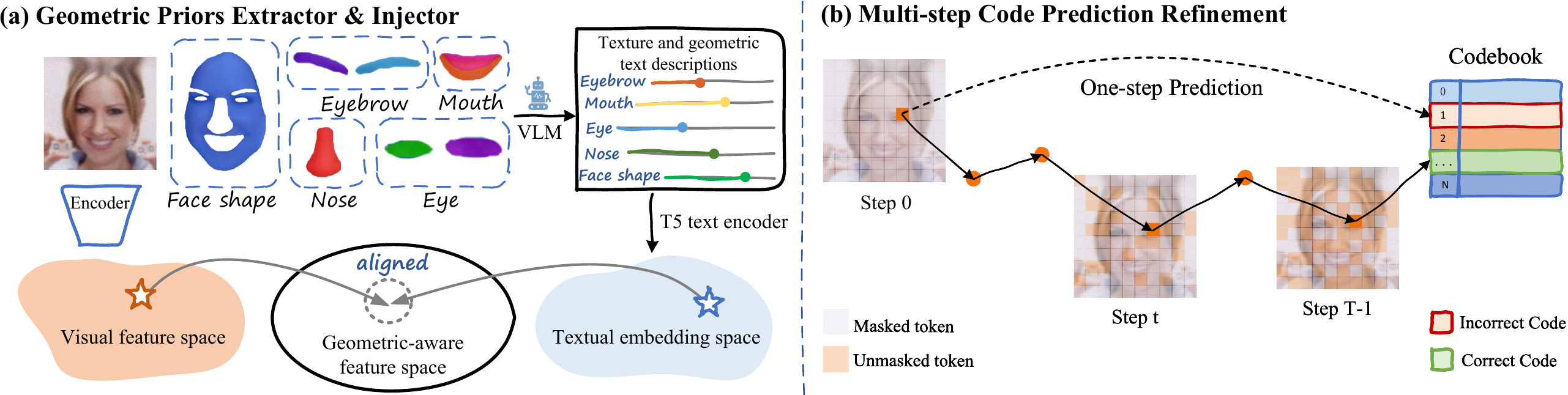}
      \caption{Our motivations are twofold: (a) aligned geometry-aware feature extraction and (b)~multi-step latent code refinement.}
    \label{Fig1}
\end{teaserfigure}


\maketitle

\section{Introduction}

Blind Face Restoration (BFR) aims to recover high-quality (HQ) images from inputs degraded by complex, unknown distortions. This task is inherently ill-posed, as multiple plausible HQ solutions can correspond to the same low-quality (LQ) observation. 

Recent codebook-based methods~\citep{gu2022vqfrblindfacerestoration, zhou2022codeformer, tsai2024dualassociatedencoderface} mitigate this ambiguity by leveraging pre-trained HQ codebooks as discrete generative priors and reformulating restoration as a latent code prediction problem. However, existing frameworks, such as CodeFormer~\citep{zhou2022codeformer} and DAEFR~\citep{tsai2024dualassociatedencoderface}, typically employ a one-step mapping strategy that predicts the entire latent code sequence from LQ features in a forward pass. This paradigm is fragile due to unreliable conditioning and limited refinement capability: severe degradation may erase component boundaries and structural evidence from the LQ representation. Ambiguous features can then be mapped to inconsistent codes across related facial regions, while the one-step predictor cannot revise uncertain structures using newly recovered context.

To overcome these limitations, we improve one-step codebook-based BFR from two synergistic perspectives: (1) enhancing the robustness of conditioning features, and (2) improving the efficacy of the prediction mechanism.

Regarding strengthening the conditioning features, recent text-conditioned generation models~\citep{rombach2022high} demonstrate strong capability in generating fine details under explicit semantic guidance. Since textual prompts are largely invariant to visual corruptions (e.g., blur, noise, and JPEG artifacts), they provide promising priors for severely degraded face restoration. Some BFR works attempt to incorporate text through paired captions or user-specified prompts~\citep{xie2023tfrgan, li2025measurement}. However, generic descriptions may not fully capture structured facial characteristics. Moreover, conventional alignment mechanisms, such as Spatial Feature Transform (SFT)~\citep{Wang2018RecoveringRT} and Cross-Attention~\citep{vaswani2017attention}, may be limited by ambiguous degraded visual features, hindering fine-grained semantic alignment. Therefore, \emph{how these textual priors are extracted and aligned with visual features becomes crucial}. Recent studies~\citep{yu2024scaling, JIANG2026105978} demonstrate the potential of vision-language models (VLMs) to extract informative semantic priors from visual inputs. To better harness this capability for structured face restoration, we propose a dual-input extraction pipeline and an Aligned Geometric Priors Injector (Fig.~\ref{Fig1} (a)). Instead of relying on the RGB image alone, we condition the VLM on both the degraded image and its parsing map, which provides explicit, spatially faithful geometric anchors. This enables more reliable component-based descriptions while reducing structural hallucinations. The injector further integrates this cross-modal guidance by modulating LQ features via SFT and then enhances them through KV-Q exchange cross-attention. By using semantically rich, prior-aligned features as queries and the original LQ features as keys and values, the injector retrieves relevant visual evidence and yields geometrically aligned representations.

Regarding improving the prediction mechanism, mapping highly corrupted LQ features to HQ codes in one step requires the model to simultaneously capture complex global structural dependencies, such as facial symmetry and jawline contours. When key structural cues are missing, this can lead to inconsistent structures, artifacts or overly smoothed outputs with limited perceptual quality. To address this, instead of conventional one-step mapping, we adapt the masked autoregressive (MAR) mechanism used in MaskGIT-based image generation~\citep{chang2022maskgitmaskedgenerativeimage} to progressively refine the predictions. In our framework, MAR reformulates latent code prediction as a multi-step, progressive refinement process, where geometrically aligned features provide reliable structural guidance. As illustrated in Fig.~\ref{Fig1} (b), instead of resolving all structural dependencies at once, MAR decomposes the complex task: it first predicts confident and structurally simple regions and then iteratively refines uncertain and complex regions (e.g., eyes and mouth) using increasingly reliable context. This coarse-to-fine, context-aware generation strategy improves structural consistency and perceptual realism by progressively masking and regenerating uncertain tokens.

In a nutshell, we propose an effective BFR framework named GeoMAR. Our main contributions are as follows:
\begin{itemize}
    \item We reformulate latent code prediction in BFR as a Masked Autoregressive process, enabling iterative context-aware refinement of facial tokens for improved structural consistency and perceptual realism.
    \item Our Geometric Priors Extractor \& Injector employs a dual-input extraction pipeline and combines Spatial Feature Transform with a KV-Q exchange Cross-Attention to learn geometrically aligned features.
    \item Extensive experiment results verify the effectiveness of our method in addressing the ambiguity of conditioning features and the fragility of the prediction mechanism.
    
\end{itemize}

\section{Related Work}

\begin{figure*}[!t]
    \centering
    \includegraphics[width=0.9\textwidth]{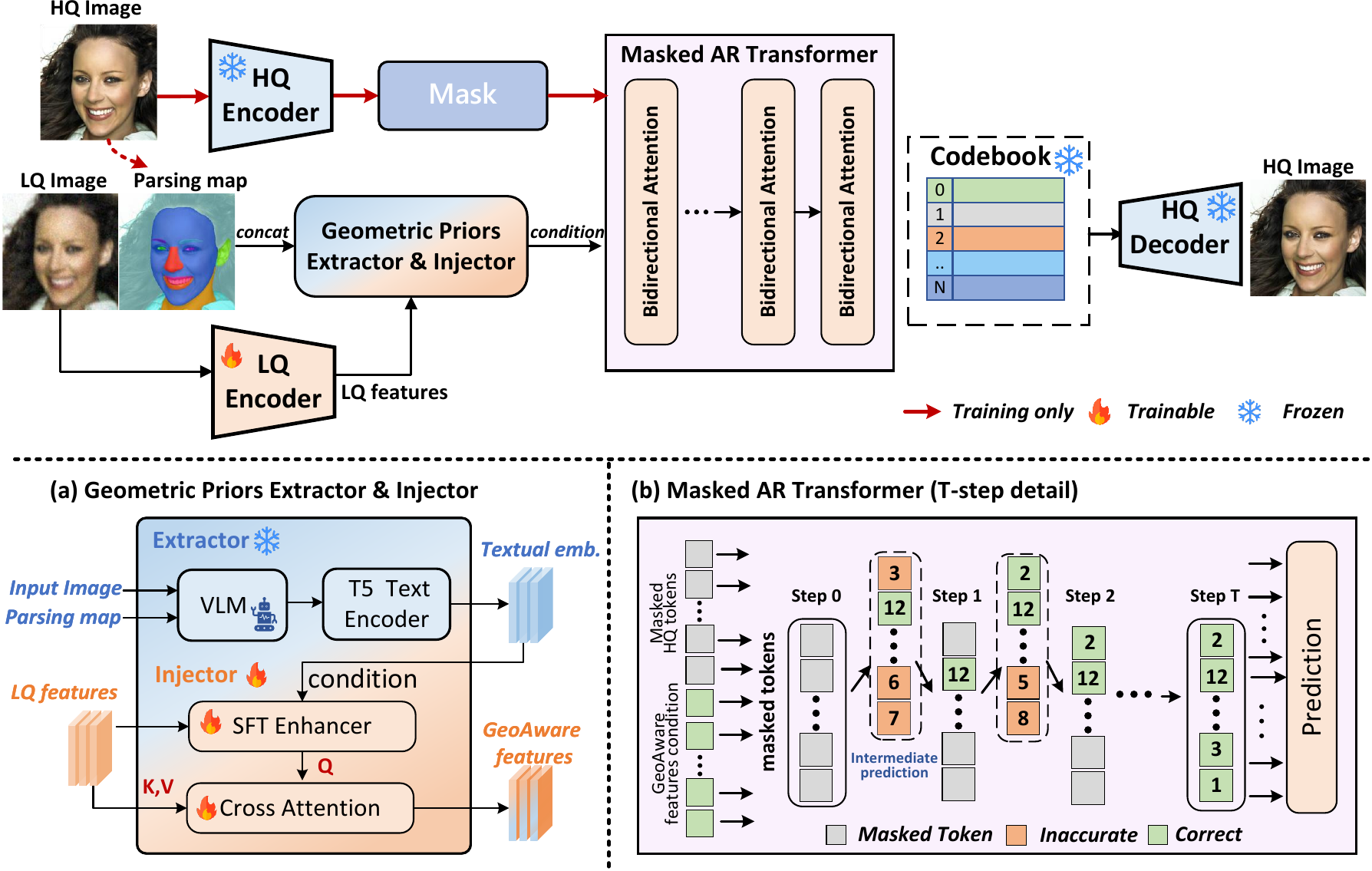}
    \caption{An overview of GeoMAR, which takes LQ images and geometry-aware facial component descriptions as inputs to restore HQ images via (a) a Geometric Priors Extractor and Injector, and (b) a Masked Autoregressive Transformer with T-step detail prediction.}
    \label{Fig2}
\end{figure*}

\paragraph{Blind face restoration.} The field of BFR has evolved through GAN-based~\citep{menon2020pulseselfsupervisedphotoupsampling,wang2021gfpgan,yang2021gpen,Chen_2021_CVPR,hu2023dear}, diffusion-based~\citep{wang2023dr2diffusionbasedrobustdegradation,lu20243d,yue2024difface,Lu_2025}, and codebook-based paradigms. Following VQ-VAE~\citep{van2017neural} and VQGAN~\citep{esser2021tamingtransformershighresolutionimage}, codebook-based methods~\citep{gu2022vqfrblindfacerestoration} exploit discrete latent priors for facial reconstruction, while RestoreFormer~\citep{wang2022restoreformer} and CodeFormer~\citep{zhou2022codeformer} employ Transformer-based architectures to predict latent codes from degraded inputs. DAEFR~\citep{tsai2024dualassociatedencoderface} further improves representation alignment via a dual-encoder design. However, these methods predominantly rely on one-step code prediction, which limits global structural consistency under severe degradation.

\paragraph{Vision-language models as image restoration priors.}
Recent image restoration methods exploit vision-language priors in different forms. SeeSR~\citep{wu2024seesr} learns degradation-aware prompt extractors from LR images, while SUPIR~\citep{yu2024scaling} and DaLPSR~\citep{JIANG2026105978} use VLM-generated descriptions to guide diffusion restoration. PURE~\citep{wei2025perceiveunderstandrestorerealworld} instead adapts an autoregressive multimodal model to predict degradation characteristics and semantic descriptions before generating HQ image tokens. In BFR, AuthFace~\citep{liang2025authface} adopts VLM-generated semantic and photography-aware tags to optimize face diffusion priors. However, these approaches mainly capture global semantics or appearance, while component-level facial geometry and its alignment with degraded features remain underexplored.

\paragraph{Autoregressive generative modeling.}
Autoregressive (AR) modeling has shown strong capability in visual generation, evolving from early pixel-level modeling approaches such as PixelRNN~\citep{oord2016pixelrecurrentneuralnetworks} and iGPT~\citep{pmlr-v119-chen20s} to more advanced Transformer-based frameworks. MaskGIT~\citep{chang2022maskgitmaskedgenerativeimage} introduces a masked token prediction paradigm with bidirectional attention, enabling parallel token generation and iterative refinement, followed by extensions such as VAR~\citep{tian2024visualautoregressivemodelingscalable} and xAR~\citep{ren2025xar}. These methods demonstrate that progressive, iterative prediction can effectively model complex dependencies and uncertainty. However, applying such masked autoregressive strategies to latent code prediction in BFR remains largely underexplored.

\section{Proposed Method}

As illustrated in Fig.~\ref{Fig2}, GeoMAR consists of a geometric priors extractor and injector and a masked autoregressive (MAR) transformer. During training, LQ images are encoded into visual features and fused with textual priors for masked latent-code prediction. In the inference stage, the framework reconstructs HQ images from LQ inputs by iteratively predicting latent codes with the MAR transformer conditioned on the geometrically aligned features.

\subsection{Preliminaries}
\label{3.1}
\paragraph{Image tokenizer.} A VQGAN-based~\citep{esser2021tamingtransformershighresolutionimage} tokenizer contains three components: an encoder, a quantizer and a decoder. Given an HQ image \(I_{HQ} \in \mathbb{R}^{H \times W \times 3}\), the HQ encoder $\mathcal{E}_H$ compresses it into latent features \(Z_{HQ} \in \mathbb{R}^{m \times n \times d}\), where $m$ and $n$ indicate the spatial size and $d$ represents the dimension of the latent vector. These features are then quantized via a pretrained codebook \(\mathcal{C} = \{c_k \in \mathbb{R}^d\}_{k=0}^{K-1}\) using the nearest neighbor lookup:
\begin{equation}
{Z_{HQ}^{c(i,j)}} = \arg\min_{c_k \in \mathcal{C}} \left\| {Z_{HQ}^{(i,j)}} - c_k \right\|_2,   k \in [0, \dots, K-1],
\end{equation}
where $Z_{HQ}^c$ is the quantized feature map of $Z_{HQ}$ with $K$ codebook entries, which is reshaped into a sequence of target discrete indices \(\boldsymbol{Y} = (y_0, \dots, y_{L-1})\), where \(L = m \cdot n\) and \(y_i \in \{0, \dots, K - 1\}\) 
represents the index of the corresponding codebook entry.

\subsection{Geometry-Aware Representation Learning}
\paragraph{Component-based geometric priors extractor.}
To obtain semantically rich and geometrically grounded priors beyond generic captions, we introduce a dual-input extraction pipeline. As illustrated in Fig.~\ref{Fig2} (a), the VLM jointly processes the degraded RGB image and parsing map to generate facial descriptions, establishing a crucial prompt-level cross-validation mechanism. Our prompt is structured to generate tag-style facial attributes by adopting a component-based descriptive logic. For each component, geometric attributes (e.g., shape and state) are derived from the parsing map, while appearance attributes (e.g., color and texture) are extracted from the RGB image. This design improves robustness in two aspects as shown in Fig.~\ref{fig:qp} and Sec.~\ref{discu}: the parsing map provides geometric cues to compensate for high-frequency information lost in LQ images, while RGB appearance cues (e.g., shadows, color gradients) help the VLM deduce a plausible description when severe degradation renders the parsing map unreliable in certain regions. We then employ Flan-T5~\citep{chung2024scaling} text encoder to extract text embeddings that capture rich semantic content as well as geometry-aware cues.

\paragraph{Aligned geometric priors injector.}

\label{sec:injector}
To effectively integrate this high-level semantic guidance with the LQ visual features, we introduce the Aligned Geometric Priors Injector, which alleviates the ambiguity of degraded visual features. As shown in Fig.~\ref{Fig2} (a), our injector uniquely combines a Spatial Feature Transform (SFT)~\citep{Wang2018RecoveringRT} and a KV-Q Exchange Strategy. Let $Z_{LQ} \in \mathbb{R}^{H \times W \times C}$ denote LQ features and $Z_{txt}$ denote the projected textual-prior embeddings encoded by the Flan-T5 model~\citep{chung2024scaling}. First, we employ SFT to semantically modulate the visual features based on the textual guidance. Specifically, the SFT branch predicts a scaling factor $s$ and a shifting factor $\beta$ from $Z_{txt}$, producing the prior-aligned features:
\begin{equation}
Z_{\mathrm{align}}=(1+s)\odot Z_{LQ}+\beta.
\end{equation}
This operation modulates LQ features by enhancing text-consistent responses and suppressing irrelevant ones, aligning feature distributions with textual priors while preserving the original spatial structure. Next, we introduce a KV-Q Exchange Strategy to fuse global semantic context with the locally-precise visual features. Unlike conventional designs that use the original LQ features as the query, we observe that unreliable queries may hinder high-fidelity restoration ~\citep{10655585}. Therefore, we use $Z_{\text{align}}$ as the query, while treating the original LQ features as the key and value: 
\begin{equation}
    Z_{enh} = \operatorname{CrossAttn}(Q=Z_{\text{align}},\, K=Z_{LQ},\, V=Z_{LQ}).
    \label{eq:cross_attention}
\end{equation}
By using the semantically rich and structurally sound $Z_{\text{align}}$ as the query, our injector can selectively retrieve the most relevant, high-fidelity information from the original LQ feature space. This guided attention process produces the final enhanced $Z_{enh}$, which are then passed to the MAR transformer.

\subsection{Masked Autoregressive Token Prediction}
To overcome the fragility of one-step prediction and improve global structural consistency, we reformulate the latent code mapping as a multi-step, progressive refinement process, where the geometrically aligned features $Z_{enh}$ are provided as a conditional prefix to the Masked AR transformer in Fig.~\ref{Fig2} (b). Specifically, let a binary mask sequence $\boldsymbol{M} = [m_{0}, \ldots, m_{L-1}] \in \{0, 1\}^L$ be applied to the target HQ code sequence $\boldsymbol{Y}$ to produce the masked sequence $\boldsymbol{Y}_M$. Here, $L$ denotes the total length of the token sequence. Elements where $m_{i} = 0$ retain their original discrete code values (acting as visible context), while positions where $m_{i} = 1$ are replaced with a special $[\mathrm{MASK}]$ token. We perform iterative refinement over $T$ steps with a dynamic masking schedule $\gamma(r) = \cos\!\left(\frac{\pi}{2} \cdot r\right)$ ($r \in (0, 1]$) following MaskGIT~\citep{chang2022maskgitmaskedgenerativeimage}, which gradually reduces the ratio of masked tokens. 

At each step $t \in [1, T]$, the transformer takes the current masked sequence $\boldsymbol{Y}^{(t)}_M$ and $Z_{enh}$ as input and predicts token probabilities $\boldsymbol{p}^{(t)}$. The confidence of previously unmasked tokens is fixed to 1.0 to preserve reliable structural context. The number of tokens to be masked in the next step is determined by $n^{(t)} = \left\lceil \gamma(t/T) \cdot L \right\rceil$. All tokens are then sorted by confidence in descending order, and the $n^{(t)}$ least confident tokens are retained as masked. The updated mask $M^{(t+1)}$ defines the next masked sequence $\boldsymbol{Y}^{(t+1)}_M$, which is progressively refined by the transformer. After \(T\) steps, the complete predicted sequence \(\hat{\boldsymbol{Y}}\) is obtained, formalized as:

\begin{equation}
P(\hat{\boldsymbol{Y}}\mid {Y}^{(1)}_M, Z_{enh}) = \prod_{t=1}^{T} P(Y^{(t+1)}_M\mid {Y}^{(t)}_M, Z_{enh}).
\end{equation}

\begin{figure*}[!t] 
    \centering 
    \includegraphics[width=1\textwidth]{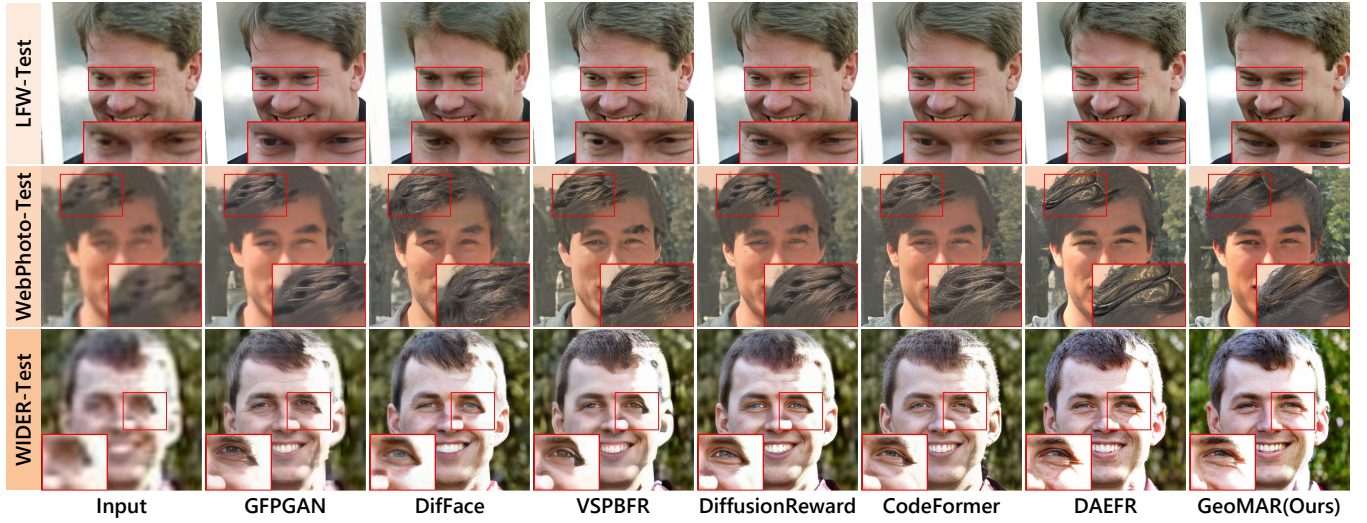} 
    \caption{Qualitative comparison on three real-world datasets: LFW-Test, WebPhoto-Test and WIDER-Test.}
    \label{result_real} 
\end{figure*}

\begin{table*}[htbp] 
\centering
\caption{Quantitative comparison on three real-world datasets. {\color{red}{Red}}, {\color{blue}{blue}} and {\color{cyan}{cyan}} indicate the best, second and third-best, respectively. For GeoMAR, $^\dagger$ denotes the online restoration-core latency; the complete end-to-end latency is described in the implementation details.}
\label{table1}

\small 
\setlength{\tabcolsep}{4pt} 

\begin{tabular}{ll c ccc ccc ccc}
    \toprule 
    \multirow{2}{*}{Prior type} & \multirow{2}{*}{Methods} & \multirow{2}{*}{Time (s)} & 
    \multicolumn{3}{c}{LFW-Test} & 
    \multicolumn{3}{c}{WebPhoto-Test} & 
    \multicolumn{3}{c}{WIDER-Test} \\
    \cmidrule(lr){4-6} \cmidrule(lr){7-9} \cmidrule(lr){10-12}
    &  && FID$\downarrow$ & NIQE$\downarrow$ & MANIQA$\uparrow$ & 
    FID$\downarrow$ & NIQE$\downarrow$ & MANIQA$\uparrow$ & 
    FID$\downarrow$ & NIQE$\downarrow$ & MANIQA$\uparrow$ \\
    \midrule 
    
    \multirow{2}{*}{GAN-based} 
    & GFPGAN [CVPR’21]&0.13& 53.86& 4.742& 0.456& 100.20& 4.970& 0.438&49.45&4.693&0.439\\
    & GPEN [CVPR’21]&0.06& 55.18& 4.538& 0.479& 85.63& 5.306& 0.438& 58.52&5.264&0.450\\
    \midrule 
    
    \multirow{4}{*}{Diffusion-based} 
    & DR2 [CVPR’23]&2.05& 50.79& 4.538& 0.525& 113.37& 6.035& 0.412& 54.70&5.904&0.407\\
    & DifFace [TPAMI’24] &7.05& \color{red}{46.35}& 3.813& 0.461& 80.69& 4.372& 0.425&37.43&4.255&0.431\\
    & VSPBFR [PR’25]&0.11& \color{blue}{46.37}& \color{cyan}{3.746}& 0.533& \color{cyan}{77.24}& \color{cyan}{4.218}& \color{blue}{0.502}& 38.20 &\color{red}{3.579}&0.493\\
    & DiffusionReward [arXiv’25]&4.23 & 48.92& 4.386& 0.517&79.49 &  5.202 & 0.483 & \color{red}{36.49} & 4.724 &0.492\\
    \midrule 
    
    \multirow{4}{*}{Codebook-based} 
    & VQFR [ECCV'22] &5.71& 50.09& 3.836&\color{cyan}{0.534}& 85.28& 4.608& 0.491&37.87&4.036&\color{cyan}{0.504}\\
    & CodeFormer [NeurIPS’22] &0.07& 52.84& 4.438& 0.527& 83.94& 5.529 & \color{red}{0.503}&38.78&4.119&0.496 \\
    & DAEFR [ICLR’24] &0.12 & 47.53 & \color{blue}{3.552}& \color{red}{0.542}& \color{blue}{75.79}& \color{blue}{3.933} & 0.493&\color{blue}{36.72}&\color{blue}{3.655}&\color{red}{0.520} \\
    & GeoMAR (Ours) &0.24$^\dagger$& \color{cyan}{46.87} & \color{red}{3.547} & \color{blue}{0.540}  & \color{red}{75.73} & \color{red}{3.930}  & \color{cyan}{0.499} &\color{cyan}{37.37} &\color{cyan}{3.762} &\color{blue}{0.510}  \\
    \bottomrule 
\end{tabular}
\end{table*}

\subsection{Training and Inference}

\paragraph{Training.} 
During training, we sample a masking ratio $r$ uniformly from $(0, 1]$ and randomly mask $\left\lceil \gamma(r) \cdot L \right\rceil$ tokens in the HQ sequence $\boldsymbol{Y}$, producing the masked sequence $\boldsymbol{Y}_M$. The enhanced geometrically aligned features $Z_{enh}$ are then concatenated with $\boldsymbol{Y}_M$ as a prefix, and the combined sequence is fed into the bidirectional transformer~\citep{devlin2019bert} for masked-code prediction. The overall training objective is defined as:
\begin{equation}
\mathcal{L} = \mathcal{L}_{contra} + \lambda_{feat} \cdot \mathcal{L}_{feat} + \mathcal{L}_{mask},
\end{equation}
where $\lambda_{feat}$ is empirically set to 10. 

First, we employ a contrastive objective to align the LQ visual features $Z_{LQ}$ with the projected textual-prior embeddings $Z_{txt}$. Let $\mathbf{z}_{LQ}^{(i)}$ and $\mathbf{z}_{txt}^{(i)}$ denote their pooled and normalized representations for the $i$-th sample. For a mini-batch of size $N$, the matched visual--text pair is treated as positive, while the remaining text representations serve as negatives. The InfoNCE-style loss~\citep{radford2021learning} is defined as:
\begin{equation}
\mathcal{L}_{contra}
=
-\frac{1}{N}\sum_{i=1}^{N}
\log
\frac{
\exp\!\left(
\mathrm{sim}(\mathbf{z}_{LQ}^{(i)},\mathbf{z}_{txt}^{(i)})/\tau
\right)
}{
\sum_{j=1}^{N}
\exp\!\left(
\mathrm{sim}(\mathbf{z}_{LQ}^{(i)},\mathbf{z}_{txt}^{(j)})/\tau
\right)
},
\end{equation}
where $\mathrm{sim}(\cdot,\cdot)$ denotes cosine similarity and $\tau$ is a learnable temperature initialized to 0.07. Second, an L2 loss aligns $Z_{enh}$ with the quantized latent codebook feature $Z_c$. This provides explicit supervision for fusing the textual priors and visual features while reducing their domain discrepancy before autoregressive prediction:
\begin{equation}
\quad
\mathcal{L}_{feat} = ||\boldsymbol{Z}_{enh} - sg(\boldsymbol{Z}_c)||_2^2,
\end{equation}
where $sg(\cdot)$ denotes the stop-gradient. Finally, $\mathcal{L}_{mask}$ minimizes the negative log-likelihood only at masked positions, conditioned on $\boldsymbol{Y}_M$ and $Z_{enh}$:
\begin{equation}
\mathcal{L}_{mask} = - \mathbb{E}_{\boldsymbol{Y}\in \mathcal{D}} \left[ \sum_{\forall i\in [1,L], m_i=1} \log p(y_i | \boldsymbol{Y}_M, \boldsymbol{Z}_{enh}) \right].
\end{equation}

\paragraph{Inference.}
The inference process differs from the training phase in two key aspects. First, without access to HQ images, we feed both the LQ image and its parsing map into the VLM to extract component-based geometric priors. This dual-input design yields priors that remain relatively robust to severe input degradation, as demonstrated in Sec.~\ref{discu}. Second, at the initial step $t=1$, the entire token sequence $Y_M^{(1)}$ is initialized with \texttt{[MASK]} tokens, allowing the model to perform blind restoration without ground truth.

\section{Experiment}

\subsection{Experimental Settings}
\paragraph{Datasets.}  We use the FFHQ dataset~\citep{karras2019style} containing 70,000 HQ images for training. All images are resized to $512\times512$. We follow~\citep{zhou2022codeformer,tsai2024dualassociatedencoderface} to synthesize corresponding LQ images. Moreover, we adopt Qwen3-VL-8B-Instruct~\citep{Qwen3-VL} to generate component-based face description prompts based on a dual-input (i.e., an RGB image and its parsing map). The parsing map is obtained using BiSeNet~\citep{yu2018bisenet}. We evaluate our method on four widely used benchmarks for BFR, including one synthetic dataset and three real-world datasets. The synthetic dataset CelebA-Test comprises 3,000 images selected from CelebA-HQ~\citep{karras2018progressivegrowinggansimproved}. To quantitatively evaluate restoration performance under different degradation, we adopt two settings: the standard GFPGAN~\citep{wang2021gfpgan} degradation pipeline (FID=144) and the degradation pipeline used during training (FID=267). The three real-world datasets, respectively, are LFW-Test~\citep{2008Labeled} (1,711 images, mild degradation), WebPhoto-Test~\citep{wang2021gfpgan} (407 images, medium degradation), and WIDER-Test~\citep{yang2016wider} (970 images, heavy degradation).

\paragraph{Evaluation metrics.} We adopt both no-reference and reference-based metrics for the comprehensive evaluation. Specifically, we use no-reference metrics FID~\citep{FID}, NIQE~\citep{NIQE}, and MANIQA~\citep{MANIQA} to evaluate the perceptual quality of all datasets. For the synthesized dataset with ground truth, we also adopt PSNR, SSIM~\citep{SSIM} and inter-ocular normalized landmark distance (LMD) for evaluation. LMD is reported in percentage in Table~\ref{tab:synthetic_comparisons}.

\paragraph{Implementation details.} In our implementation, the size of the input face image is $512\times512\times3$, and the size of the quantized feature is $16\times16\times256$. The pre-trained codebook~\citep{tsai2024dualassociatedencoderface} contains $K = 1,024$ code items, with each code represented by a 256-dimensional vector. During training, we employ the Adam optimizer~\citep{kingma2015adam} with a batch size 16 and set the learning rate to $7.2\times10^{-5}$. In autoregressive token prediction, the refinement step is empirically set to 8. The model is trained with four NVIDIA GeForce RTX 4090 GPUs for 28 epochs. During inference, the complete pipeline takes around $3.91$ seconds per image for a newly observed input, comprising face parsing with BiSeNet ($0.03$ s), prompt generation with Qwen3-VL-8B-Instruct ($3.63$ s), text encoding with Flan-T5 ($0.01$ s), and MAR prediction plus decoding with the GeoMAR core ($0.24$ s$^\dagger$ as reported in Table~\ref{table1}). 

\begin{figure*}[!t]
\centering
    \includegraphics[width=1\textwidth]{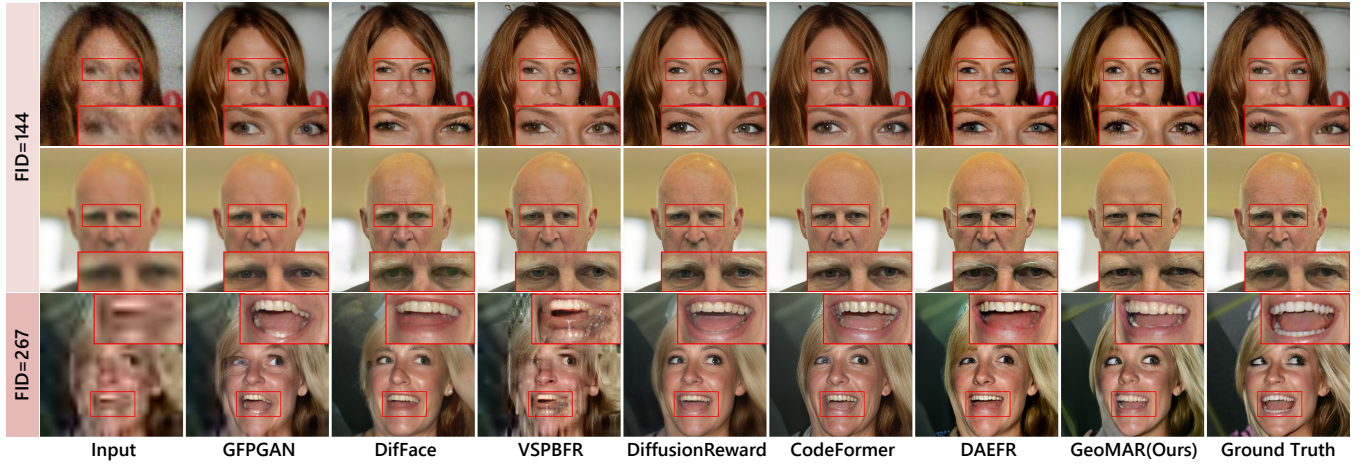}

    \caption{Qualitative comparison on the synthetic CelebA-Test dataset under two degradation levels (FID=144 and FID=267).}
    \label{celeba_compare}
\end{figure*}

\begin{table*}[t]
\centering
\caption{Quantitative comparison on the synthetic CelebA-Test dataset under two degradation levels (FID=144 and FID=267). 
{\color{red}{Red}}, {\color{blue}{blue}} and {\color{cyan}{cyan}} indicate the best, second and third-best, respectively.}
\label{tab:synthetic_comparisons}

\small 
\setlength{\tabcolsep}{3pt} 

\begin{tabular}{ll cccccc cccccc}
    \toprule 
    \multirow{2}{*}{Prior type} & \multirow{2}{*}{Methods} & \multicolumn{6}{c}{FID = 144} & \multicolumn{6}{c}{FID = 267} \\
    \cmidrule(lr){3-8} \cmidrule(lr){9-14}
    & & FID$\downarrow$ & NIQE$\downarrow$ & MANIQA$\uparrow$ & PSNR$\uparrow$ & SSIM$\uparrow$& LMD $\downarrow$& FID$\downarrow$ & NIQE$\downarrow$ & MANIQA$\uparrow$ & PSNR$\uparrow$ & SSIM$\uparrow$ & LMD $\downarrow$\\
    \midrule

    \multirow{2}{*}{GAN-based} 
    & GFPGAN [CVPR’21]&52.29&4.565&0.466&\color{blue}{25.80}&\color{red}{0.708}& 2.52 &79.30& 5.451 & 0.453 &  21.97 & 0.611& 8.89 \\
    & GPEN [CVPR’21]& 51.46&4.817&0.459& \color{cyan}{25.77} &\color{blue}{0.703}&2.50& 72.99 & 5.723 & 0.447 & 22.05 &  \color{cyan}{0.619}&9.19   \\
    \midrule 
    
    \multirow{4}{*}{Diffusion-based}
    & DR2 [CVPR’23]& 58.59&5.370&0.487&24.45&0.679& 2.66& 72.41 & 5.579 & 0.477 & 21.28 & 0.606& 4.40 \\
    & DifFace [TPAMI’24]& \color{red}{37.88}&4.157&0.468&24.80&0.673&  2.47& 47.89 & 4.513 & 0.461 & \color{blue}{22.81} & 0.604 &\color{blue}{ 3.16} \\
    & VSPBFR [PR’25]& \color{blue}{38.52}&\color{cyan}{3.839}&\color{red}{0.546}&24.73 &0.657&\color{cyan}{ 2.37} & \color{blue}{43.28}& \color{red}{3.512} & 0.500 & 21.02& 0.503& 4.76    \\
    & DiffusionReward [arXiv’25] & 42.59   &4.545  &0.537 &\color{red}{26.33} &\color{cyan}{0.701}&\color{red}{ 2.15}&49.70 &5.021&0.534&\color{red}{23.25}& \color{red}{0.640}&\color{red}{ 3.04} \\
    \midrule 
    
    \multirow{4}{*}{Codebook-based}
    & VQFR [ECCV'22]&46.77 &4.204& 0.540 &23.76 &0.659& 2.45& 53.73 & 4.743 & 0.553 & 21.84 & \color{blue}{0.621}& 3.28   \\
    & CodeFormer [NeurIPS’22]& 52.42&4.594&0.538&25.15&0.670&\color{blue}{ 2.32 } &58.31 & 4.566 & \color{blue}{0.556} &\color{cyan}{22.19} & 0.594 &   \color{cyan}{3.18}  \\
    & DAEFR [ICLR’24]& 40.76&\color{red}{3.744}&\color{blue}{0.545}&22.59 &0.609 & 2.44    &\color{cyan}{47.50} &\color{cyan}{4.500} & \color{red}{0.570} & 20.29 & 0.588&3.29  \\
    & GeoMAR (Ours) &\color{cyan}{39.61}&\color{blue}{3.819} &\color{cyan}{0.541}&  21.95 &0.597& 2.66 &\color{red}{42.13} &\color{blue}{4.223} &\color{cyan}{0.554}&20.23&0.562& 3.62     \\
    \bottomrule 
\end{tabular}
\end{table*}

\subsection{Main Results}
We compare GeoMAR with nine representative BFR approaches, including: (1) GAN-based methods: GFPGAN~\citep{wang2021gfpgan}, GPEN~\citep{yang2021gpen}; (2) Diffusion-based methods: DR2~\citep{wang2023dr2diffusionbasedrobustdegradation}, DifFace~\citep{yue2024difface}, VSPBFR~\citep{Lu_2025}, DiffusionReward~\citep{wu2025diffusionrewardenhancingblindface}; (3) Codebook-based methods: VQFR~\citep{gu2022vqfrblindfacerestoration}, CodeFormer~\citep{zhou2022codeformer}, DAEFR~\citep{tsai2024dualassociatedencoderface}.

\paragraph{Evaluation on three real-world datasets.}
As shown in Table~\ref{table1}, our method consistently improves perceptual quality while maintaining practical inference time. While some methods may achieve marginally better scores on certain metrics by producing over-smoothed outputs, our method prioritizes global structural consistency and perceptual realism. Consequently, GeoMAR achieves the lowest FID on WebPhoto-Test (75.73) and the best NIQE on both LFW-Test (3.547) and WebPhoto-Test (3.930), while consistently ranking among the top three in MANIQA across all datasets, demonstrating its strong perceptual quality, as MANIQA is well aligned with human perception.

This quantitative advantage aligns with the visual comparisons in Fig.~\ref{result_real}. Under real-world degradation, many GAN-based and diffusion-based methods fail to recover complex facial details (e.g., pupils in the first and third rows) and produce over-smoothed results with artifacts, while codebook-based methods such as CodeFormer and DAEFR often yield distorted and unnatural structures due to the fragility of one-step mapping. In contrast, by combining explicit geometry-aware priors with a progressive MAR refinement process, GeoMAR reliably restores coherent facial structures and fine-grained textures (e.g., hair in the second row), producing restorations that are structurally accurate and faithful to the original attributes.

\paragraph{Evaluation on the synthetic CelebA-Test dataset.}
As shown in Table~\ref{tab:synthetic_comparisons}, GeoMAR maintains highly competitive performance under different synthetic degradation levels, ranking among the top three across all perceptual metrics under varying degradation levels. Although GeoMAR is not consistently superior across all quantitative metrics, it better preserves perceptual quality and structural realism under severe degradation. In such cases, slight shifts in facial components (e.g., eyes and mouth) may increase LMD despite visually coherent restorations.

The qualitative comparisons in Fig.~\ref{celeba_compare} further illustrate this distinction. Guided by reliable geometric anchors and refined through context-aware MAR generation, GeoMAR reconstructs complex, attribute-consistent details around the eyes in the first two rows. In the third row, GeoMAR better preserves the subject's overall facial appearance, whereas competing methods produce noticeable artifacts or appear less faithful to the original subject.

\begin{table}[t]
\centering
\caption{Ablation study of GeoMAR design on CelebA-Test dataset~(FID=144).}

\label{ab_components} 
\setlength{\tabcolsep}{8pt} 

\begin{tabular}{c cc cc}
    \toprule
    Models & Geometric Priors & MAR & FID$\downarrow$ & NIQE$\downarrow$ \\
    \midrule
    (a)  & \xmark & \xmark & 42.38 & 3.975 \\
    (b)  & \checkmark & \xmark & 42.23 & 3.891 \\
    (c)  & \xmark & \checkmark & 42.36 & 3.829 \\
    \midrule 
    (d) & \checkmark & \checkmark & \textcolor{red}{39.61} & \textcolor{red}{3.819} \\
    \bottomrule
\end{tabular}
\end{table}

\begin{table}[t]
\centering
\caption{Ablation study on geometric priors design.}
\label{tab:ablation_prompt}

\setlength{\tabcolsep}{3pt} 

\begin{tabular}{l ccc ccc}
    \toprule
    
    \multirow{3}{*}{Metrics} 
    & \multicolumn{3}{c}{LFW-Test} 
    & \multicolumn{3}{c}{WebPhoto-Test} \\
    \cmidrule(lr){2-4} \cmidrule(lr){5-7} 
    
    & \multicolumn{3}{c}{Geometric priors} 
    & \multicolumn{3}{c}{Geometric priors} \\
    
    & w/o & concat-style & Ours 
    & w/o & concat-style & Ours \\
    \midrule
    
    FID$\downarrow$ & 50.98  & 48.43  & \textcolor{red}{46.87} 
    & 75.85 & 77.74 & \textcolor{red}{75.73} \\
    NIQE$\downarrow$ & \textcolor{red}{3.496} & 3.518 & 3.547
    & 4.069 & 4.047  & \textcolor{red}{3.930} \\
    \bottomrule
\end{tabular}
\end{table}

\begin{figure}[!t]
    \centering

        \includegraphics[width=0.97\linewidth]{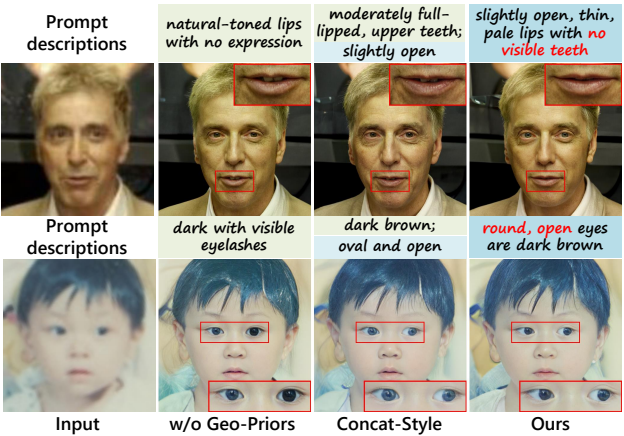}

    \caption{Qualitative comparison of different textual priors and their corresponding visual restoration results.}
    \label{geo_ab}
\end{figure}

\subsection{Ablation Studies}
\paragraph{Effectiveness of GeoMAR design.}
As shown in Table~\ref{ab_components}, the baseline model (a), which adopts one-step code mapping with generic caption-style text priors, achieves the poorest performance. Introducing explicit geometric priors (model b) improves the results, demonstrating that explicit, fine-grained structural guidance helps reduce the ambiguity in degraded LQ features and better preserves facial fidelity. Replacing one-step mapping with the masked autoregressive (MAR) mechanism (model c) consistently further boosts metrics, underscoring the value of multi-step, context-aware refinement. Crucially, our full GeoMAR model (d) integrates both innovations, exhibiting strong synergy that substantially reduces FID from 42.38 to 39.61 and NIQE from 3.975 to 3.819. These results verify their complementarity: the geometrically aligned features indeed provide a more accurate and structurally sound foundation, which the iterative, multi-step mapping of the MAR mechanism establishes global structural consistency and perceptual realism.

\paragraph{Advantage of leveraging geometric priors.}

We compare component-based geometric prior with two alternatives under the same injector architecture: (1) generic caption-style textual descriptions solely from RGB inputs (w/o Geo-Priors), and (2) a concat-style variant that generates texture and geometry descriptions separately from the image and parsing map before concatenation. As shown in Table~\ref{tab:ablation_prompt}, our design achieves the best FID. Although the ‘w/o Geo-Priors’ variant obtains slightly better NIQE on LFW-Test, this is likely due to NIQE favoring smoother outputs. By jointly leveraging RGB image and parsing map and generating descriptions based on components, the VLM forms a coherent representation with a robust cross-validation mechanism. The parsing map provides high-frequency structural guidance, while the LQ image compensates for potential parsing inaccuracies, improving robustness under severe degradation. As illustrated in Fig.~\ref{geo_ab}, relying solely on RGB inputs (w/o Geo-Priors) produces generic or structurally ambiguous results (e.g., misaligned lip contours). The concat-style variant, lacking overall context, often introduces disjointed artifacts. In contrast, our design achieves more accurate and stable restorations.

\begin{table}[t]
\centering
\caption{Ablation of query types in geometric prior injector.}
\label{tab:ablation_kv}
\footnotesize
\setlength{\tabcolsep}{4pt} 

\begin{tabular}{l cccc} 
    \toprule
    \multirow{2}{*}{Metrics} & \multicolumn{2}{c}{LFW-Test} & \multicolumn{2}{c}{WebPhoto-Test} \\
    \cmidrule(lr){2-3} \cmidrule(lr){4-5} 

    & LQ fea as \textit{Q} & Enhanced fea as \textit{Q} 
    & LQ fea as \textit{Q} & Enhanced fea as \textit{Q} \\
    \midrule
    
    FID$\downarrow$  & 48.05 & \textcolor{red}{46.87} & 76.68 & \textcolor{red}{75.73} \\
    NIQE$\downarrow$ & 3.671 & \textcolor{red}{3.547} & 4.175 & \textcolor{red}{3.930} \\
    \bottomrule
\end{tabular}
\end{table}

\begin{figure}[!t]
    \centering

        \includegraphics[width=0.85\linewidth]{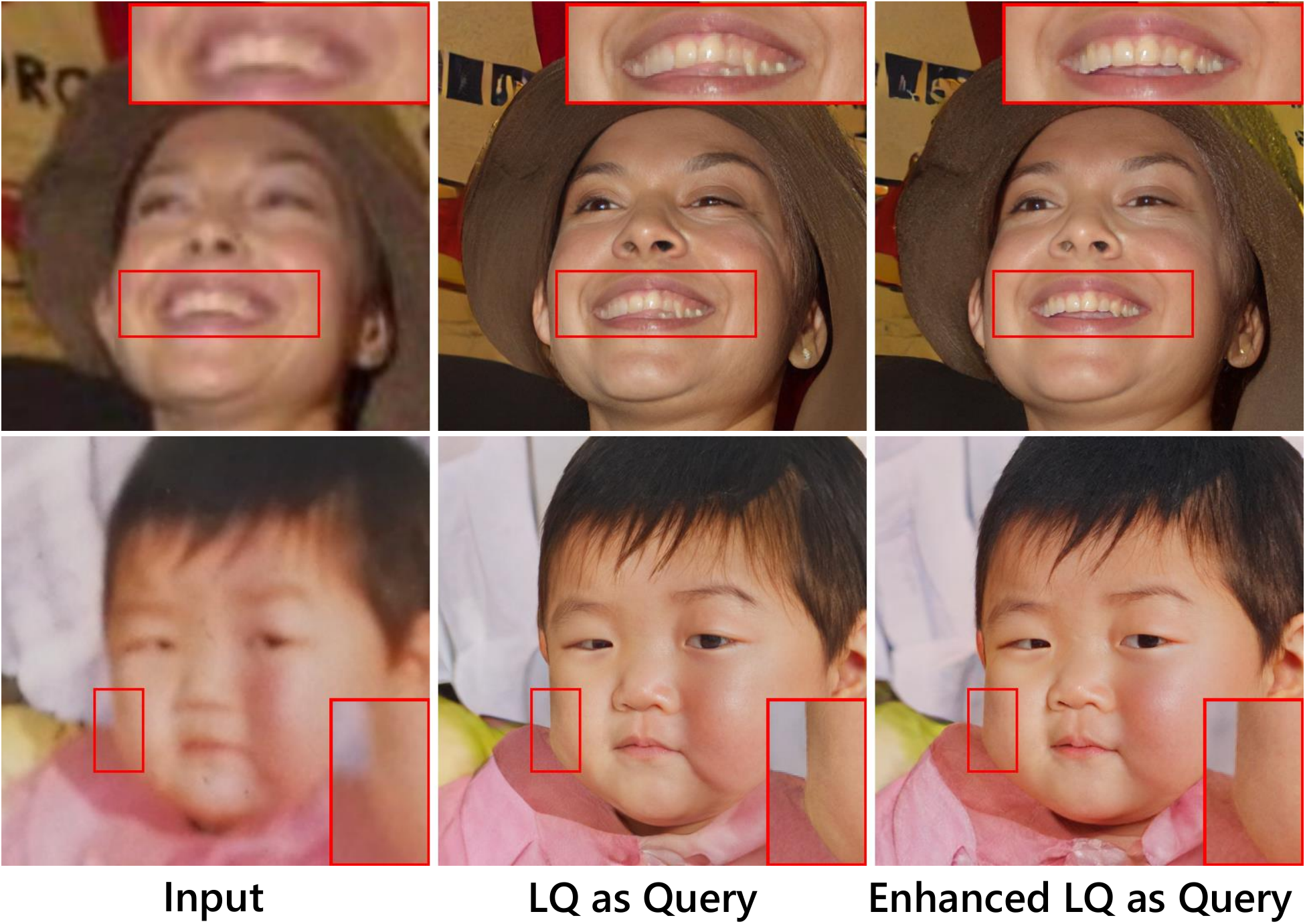}
   
    \caption{Qualitative comparison of different query designs.}
    \label{kqv}
\end{figure}

\paragraph{Insights of KV-Q exchange in geometric prior injector.}
To determine the most effective way to inject our geometric priors, we analyze the design of our Aligned Geometric Priors Injector about the role of the query in its cross-attention mechanism. We compare two feature types as queries: LQ and {Enhanced LQ} features. As shown in Table~\ref{tab:ablation_kv}, using the {Enhanced LQ} features as the query outperforms using LQ features. When the degraded LQ features serve as the query, their limited semantic information makes them vulnerable to being dominated by textual priors, causing the generic structure of the prior to override input-specific evidence. In contrast, {Enhanced LQ} features enable the prior-aligned representation to selectively retrieve relevant information from the original LQ space, allowing the geometric prior to correct global facial structure while retaining facial attributes. Fig.~\ref{kqv} provides consistent qualitative evidence: the KV-Q exchange design reduces visual ambiguity, successfully retaining realistic facial attributes (e.g., natural teeth structure and facial contour) while preventing the text from dominating the actual visual input.

\paragraph{Generalization of masked autoregressive.}

\begin{table}[t]
\centering
\caption{Generalization of MAR across BFR methods (FID). $\triangle$ indicates performance improvement.}
\label{tab:ablation_rm}
\setlength{\tabcolsep}{4pt} 

\begin{tabular}{ll cc} 
    \toprule
    Models & Methods & LFW-Test & WebPhoto-Test \\
    \midrule
    
    \multirow{2}{*}{DAEFR} 
      & w/o MAR & 47.69 & 75.79 \\
      & w/ MAR  & \textcolor{red}{45.77} ($\triangle$1.92) 
                & \textcolor{red}{73.13} ($\triangle$2.66) \\
    \midrule 
    
    \multirow{2}{*}{GeoMAR (Ours)} 
      & w/o MAR & 53.42 & 76.81 \\
      & w/ MAR  & \textcolor{red}{46.87} ($\triangle$6.55) 
                & \textcolor{red}{75.73} ($\triangle$1.08) \\
    \bottomrule
\end{tabular}
\end{table}

\begin{figure}[t!]
    \centering

        \includegraphics[width=0.97\linewidth]{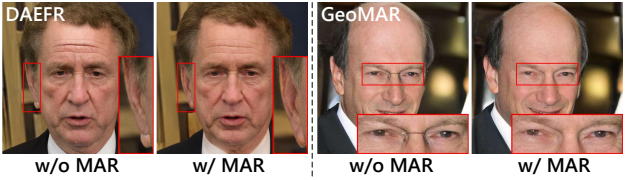}
    \caption{Generalization of MAR. MAR improves restoration quality on both GeoMAR and DAEFR.}
    \label{mar_ab}
\end{figure}

To evaluate the generalization of MAR mechanism, we integrate it as a plug-and-play module into the competitive codebook-based method DAEFR. As shown in Table~\ref{tab:ablation_rm}, MAR consistently improves performance across different architectures, demonstrating its strong generalization and its ability to alleviate the inherent fragility of one-step mapping. Qualitative results in Fig.~\ref{mar_ab} further highlight this advantage. Without MAR, DAEFR and GeoMAR predict the complete code sequence in one pass and struggle to capture complex global structural dependencies, often generating structurally inconsistent artifacts around challenging regions like eyeglasses and facial contours. In contrast, MAR reformulates the prediction into a multi-step, context-aware refinement. By establishing a reliable global structure first, MAR progressively synthesizes clean, coherent boundaries, thereby guaranteeing better structural consistency and perceptual realism.

\paragraph{Effect of prediction steps $T$ in masked autoregressive.}

Table~\ref{tab:ablation_t} reports the impact of prediction steps $T$ in MAR. Increasing $T$ consistently improves restoration quality during the early stages, as the model is able to progressively refine uncertain or complex facial regions. When $T$ increases from 1 to 8, both FID and NIQE decrease steadily on LFW-Test and WebPhoto-Test, indicating that multi-step refinement effectively reduces structural ambiguity far better than one-step mapping. However, further increasing $T$ brings diminishing returns while increasing inference cost. Therefore, we set $T=8$ as the default configuration, achieving a favorable balance between restoration quality and efficiency.

\begin{table}[t]
\centering
\caption{Quantitative results with different prediction steps $T$ in masked autoregressive.}
\label{tab:ablation_t}
\setlength{\tabcolsep}{7pt}

\begin{tabular}{c ccccc}
    \toprule
    \multirow{2}{*}{Steps $T$} & \multirow{2}{*}{Time (s)} 
    & \multicolumn{2}{c}{LFW-Test} 
    & \multicolumn{2}{c}{WebPhoto-Test} \\
    
    \cmidrule(lr){3-4} \cmidrule(lr){5-6}
    
    & 
    & FID$\downarrow$ & NIQE$\downarrow$ 
    & FID$\downarrow$ & NIQE$\downarrow$ \\
    
    \midrule
    
    1  & 0.12 & 48.24 & 3.569 & 77.23 & 3.966 \\
    4  & 0.18                  & 46.99 & 3.554 & 76.03 & 3.955 \\
    8  & 0.24                  & \textcolor{red}{46.87} & \textcolor{red}{3.547} 
                              & \textcolor{red}{75.73} & \textcolor{red}{3.930} \\
    12 & 0.32                  & 46.97 & 3.557 & 75.95 & 3.936 \\
    
    \bottomrule
\end{tabular}
\end{table}

\subsection{Discussion}
\label{discu}

\begin{figure}[ht]
    \centering

    \begin{subfigure}[t]{0.36\linewidth}
        \centering
        \vspace{0pt}
        \includegraphics[width=\linewidth]{Figures/fig7.pdf}
    \end{subfigure}
    \hfill
    \begin{subfigure}[t]{0.62\linewidth}
        \centering
        \vspace{0pt}
        \includegraphics[width=\linewidth]{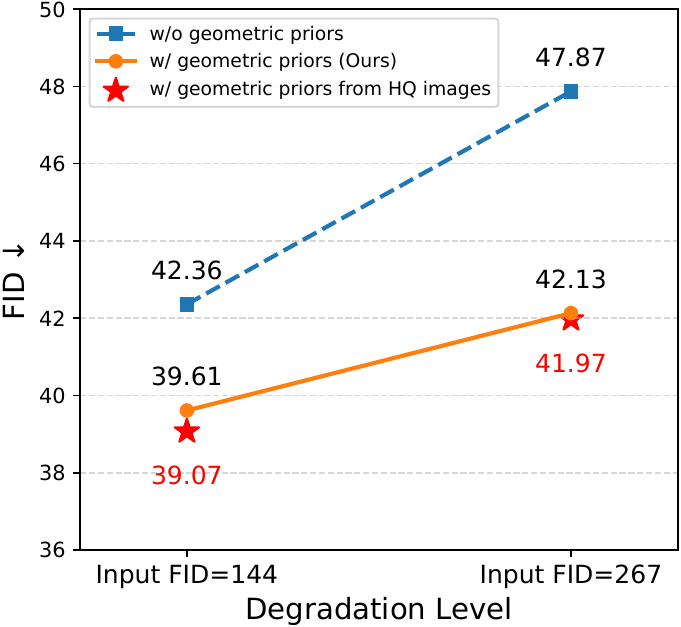}
    \end{subfigure}
    
    \caption{Comparison with and without geometric priors under different degradation levels on CelebA-Test dataset.}
    \label{fig:qp}
\end{figure}
\paragraph{Robustness of GeoMAR using inaccurate parsing maps.}
The fragility of geometric priors lies in the fact that severe input degradation may lead to inaccurate face parsing maps. To quantify its impact, we evaluate GeoMAR on CelebA-Test dataset under two degradation levels. As shown in Fig.~\ref{fig:qp} (Right), GeoMAR remains markedly more stable than the baseline without geometric priors as degradation increases. Moreover, GeoMAR stays substantially closer to the oracle setting that uses GT parsing maps (denoted by \textcolor{red}{$\star$}), indicating that our geometrically aligned features provide rather robust and reliable guidance even when the predicted parsing maps are inaccurate. Qualitative results in Fig.~\ref{fig:qp} (Left) further confirm that even when the degradation is severe, GeoMAR preserves plausible structures around challenging regions like eyes and mouths.

\section{Limitations}
\label{sec:limitations}
GeoMAR assumes that degraded RGB inputs and predicted parsing maps preserve sufficient evidence for component-level facial geometry. This assumption may fail under extreme real-world occlusions (e.g., sunglasses, hands, or foreground objects), where both RGB evidence and parsing results become unreliable, weakening the dual-input validation mechanism and causing inaccurate VLM descriptions. Consequently, GeoMAR may generate plausible but input-inconsistent details in heavily occluded regions, with failure cases presented in the supplementary material. Future work will explore occlusion-aware parsing, confidence-guided geometric prior selection, and targeted occlusion augmentation. 

\section{Conclusion}
In this paper, we present GeoMAR, a novel framework that addresses the limitations of ambiguous conditioning features and a fragile prediction mechanism in one-step codebook-based blind face restoration. Through a dual-input extraction pipeline and KV-Q Exchange Strategy, GeoMAR produces geometrically aligned features with reliable visual guidance. Furthermore, MAR-based latent code refinement enables coarse-to-fine, context-aware generation with improved structural consistency and perceptual realism. Extensive experiments demonstrate that GeoMAR achieves competitive performance against existing methods in both quantitative and qualitative evaluations.

\clearpage
\newpage
\begin{acks}
This work was supported by the National Natural Science Foundation of China (Grant No. 62206123) and the Key Science and Technology Program of Lhasa Municipality (Grant No. LSKJ202612).
\end{acks}

\bibliographystyle{ACM-Reference-Format}
\balance
\bibliography{sample-base}

\clearpage
\newpage
\appendix

\section{Overview}

This supplementary material provides additional details and results to support the findings presented in the main paper. It consists of four parts: 
\begin{itemize}
\item  More Details on GeoMAR (Section 1): We elaborate on the implementation specifics of our method, including the degradation synthesis pipeline, the design of VLM prompts for geometric priors extraction, and detailed network architectures. In addition, we present a thorough analysis of model complexity and inference efficiency.
  \item  More Discussions on GeoMAR (Section 2): We further study robustness to imperfect geometric priors, generalization under extreme real-world degradations, fine-grained component contributions, sensitivity to the choice of VLM, the perception--distortion trade-off, and the refinement mechanism of MAR.
\item  More Results on GeoMAR (Section 3): We present more qualitative comparisons to validate the effectiveness of our core components, along with additional visual results compared to state-of-the-art methods. 
\item  Limitations (Section 4): The limitations of GeoMAR are given in this section.
\end{itemize}


\section{More Details on GeoMAR}
\label{sec2}

\subsection{Details of Degradation Pipeline}
We train our model on the FFHQ dataset, which contains 70,000 high-quality face images. All images are resized to $512 \times 512$ for training. The paired degraded images are synthesized following the same procedure as in the compared methods, CodeFormer and DAEFR, which can be formulated as follows:
\begin{equation}
    I_{\text{L}} = \left\{ \left[ (I_{\text{H}} \otimes k_\sigma) \downarrow_r + n_\delta \right] \text{JPEG}_q \right\} \uparrow_r.
\end{equation}
First, the HQ image $I_{\text{H}}$ is blurred by convolution operator $\otimes$ with a Gaussian kernel $k_\sigma$. Second, a downsampling operation $\downarrow$ with a scaling factor $r$ is performed to reduce the resolution of image. Next, additive Gaussian noise $n_\delta$ is added to the downsampled image. JPEG compression with a quality factor $q$ is applied to further degrade image quality. The resulting image is then upsampled $\uparrow$ with a scaling factor $r$ to $512 \times 512$ to obtain the degraded $I_{\text{L}}$ image. In our experiment setting, we randomly sample $\sigma, r, \delta,$ and $q$ from $[0.1, 15], [0.8, 30], [0, 20],$ and $[30, 100]$, respectively. 

\begin{table*}[t]
\centering
\footnotesize
\setlength{\tabcolsep}{4.5pt}
\renewcommand{\arraystretch}{1.1}
\caption{System prompt template and a component-based example used in the VLM pipeline.}
\label{tab:vlm_prompt}
\begin{tabular}{p{2.8cm}p{14.3cm}}
\toprule
\textbf{Item} & \textbf{Content} \\
\midrule
\textbf{Role} & ``You are an expert visual analyst for high-fidelity face restoration, specialized in generating dense natural-language descriptions for T5 text encoders.'' \\
\midrule
\textbf{Input definition} & ``Image 1 (RGB): source for color, texture, lighting, and fine details. Image 2 (Parsing Map): source for geometry, including shapes, open/closed states, and relative sizes.'' \\
\midrule
\textbf{Core rules} & ``Generate a single coherent and objective subsection. Do not use subjective metaphors. Start from global appearance and lighting, then describe Face $\rightarrow$ Eyes/Brows $\rightarrow$ Nose $\rightarrow$ Mouth. Ignore the colors in the parsing map itself. Use the parsing map only for geometry and use the RGB image for actual appearance.'' \\
\midrule
\textbf{Component templates} & ``Overall appearance \& accessories:'' age, gender, race, skin tone, hair, pose, expression, accessories, and lighting from RGB. ``Face \& skin details:'' face shape from parsing map, but freckles/makeup/wrinkles from RGB. ``Eyes \& eyebrows:'' eye shape/state from parsing map, iris color and lashes from RGB. ``Nose \& mouth:'' shape/size/state from parsing map, lip texture/color and teeth visibility from RGB. \\
\midrule
\textbf{Output format} & ``Return only the raw subsection text. Do not output tags, section headers, or bullet points.'' \\
\midrule
\textbf{Example output} & ``A young Asian female with pale skin and long black hair with bangs is smiling under soft natural lighting. She is wearing circular gold earrings and a silver necklace. She features an oval face shape with a heavy makeup finish and no visible freckles. Her narrow, open eyes are dark brown with thick eyelashes, complemented by arched, neatly groomed black eyebrows. She has a small nose with smooth skin texture. Her mouth is slightly open, displaying thick, glossy red lips with upper teeth visible.'' \\
\bottomrule
\end{tabular}
\end{table*}

\subsection{Details of Geometric Priors Extraction Pipeline}
\begin{figure} 
    \centering 
    \includegraphics[width=0.5\textwidth]{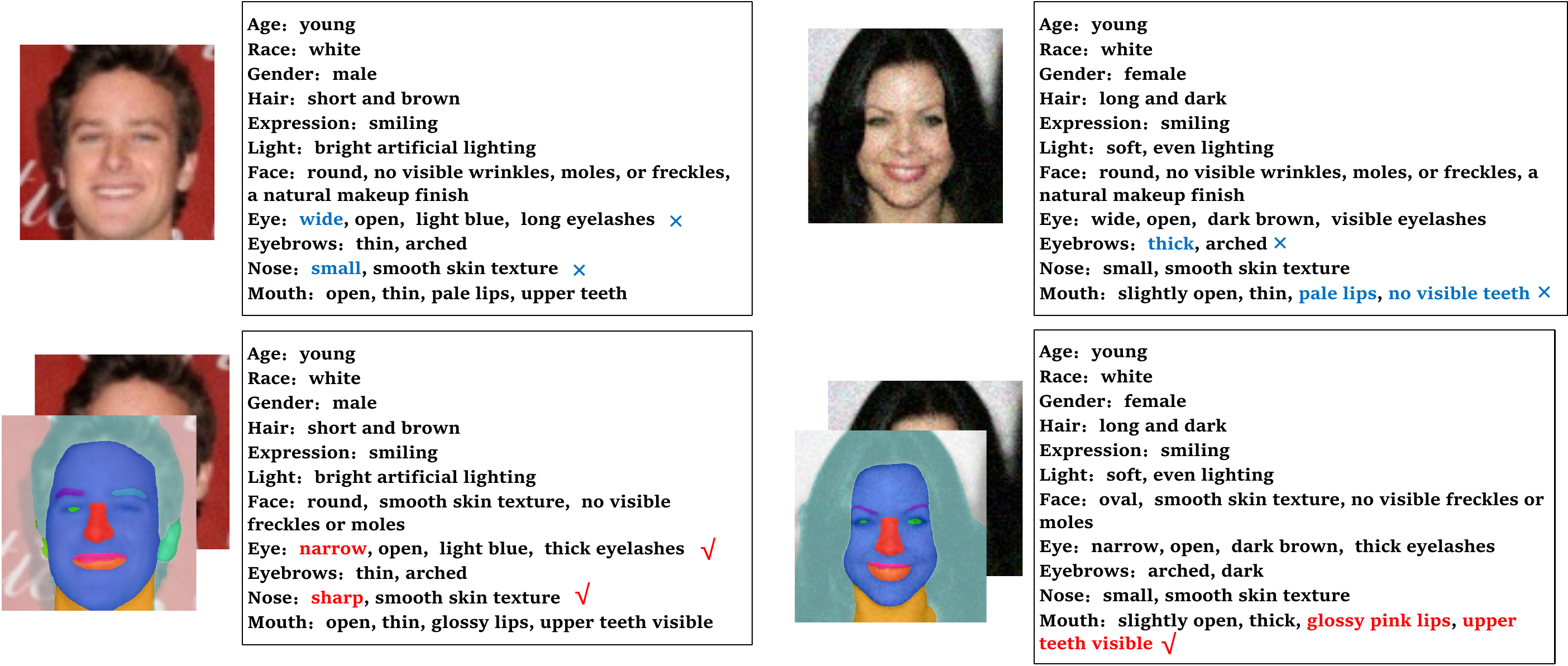} 
    \caption{Comparison of the tag description accuracy between original images and parsing maps.}
    \label{prompt} 
\end{figure}

To obtain geometric priors, we first predefine a structured set of facial attributes that explicitly cover both appearance-level details and geometry-related properties. We instantiate the VLM with Qwen3-VL-8B-Instruct, i.e., a 8B-parameter multimodal chat model. Instead of concatenating independent captions, we feed the VLM with a two-image input tuple in a fixed order: Image 1 is the LQ RGB face image and Image 2 is the corresponding facial parsing map. In implementation, the two inputs are packed into the same chat turn as two image entries rather than concatenated along the channel dimension. As shown in Table~\ref{tab:vlm_prompt}, the prompt enforces a component-wise reasoning process, where geometric attributes (e.g., shape and open/closed states) are inferred only from the parsing map, while appearance-related attributes (e.g., color, texture, lighting, and accessories) are inferred from the RGB image. This design improves robustness under severe degradations: when the parsing map is locally unreliable, the RGB image provides complementary contextual cues for correction. The VLM outputs a single coherent subsection that is semantically consistent and plausible, thereby reducing hallucinated descriptions that contradict facial structure. We then encode this subsection with a frozen T5-Large encoder into a sequence of 128 tokens, each represented by a 1024-dimensional embedding, and use it as the textual prior for GeoMAR. Importantly, both the VLM and T5 are used only in an offline pre-processing stage to extract and save text features; during GeoMAR training and inference, we only load the pre-computed T5 features and no gradient is propagated to either model. As shown in Fig~\ref{prompt}, the prompts explicitly emphasize critical facial components and provide more accurate structural cues than conventional semantic captions.

\subsection{Details of GeoMAR's Architecture}
The architecture of GeoMAR consists of a hierarchical encoder-decoder design and a masked auto-regressive transformer for reconstruction. Both HQ and LQ encoders share the same structural design, ensuring consistency across feature spaces. During training, the HQ codebook, decoder, post-quantization convolution, HQ encoder, and HQ quantization convolution are frozen, while only the LQ encoder is updated to map degraded inputs into the frozen HQ token space.

For masked auto-regressive prediction, we employ a transformer with nine stacked self-attention blocks, embedding dimension 512, 8 attention heads, and an MLP width of 1024 per block. The aligned LQ feature sequence $L_E \in \mathbb{R}^{256 \times 256}$ is linearly projected from 256 to 512 channels and concatenated with the masked HQ token embeddings, forming the transformer input of 512 tokens (256 LQ prefix tokens + 256 HQ target tokens). Long-range skip connections are fused by concatenating feature maps from the main and skip branches ($\mathbf{h}_m$ and $\mathbf{h}s$) along the channel dimension and applying a linear projection: $h{fused} = Linear(Concat(h_m, h_s))$, enabling reconstruction guidance from $L_E$ while preserving the flexibility of iterative masked token prediction.

The Injector aligns textual and visual features through a four-stage module. First, frozen T5-Large features of shape $128\times1024$ are projected to $128\times256$ via a learnable linear projector with GELU and LayerNorm. Second, a 3-layer text MLP ($256 \rightarrow 512 \rightarrow 512 \rightarrow 256$) refines the sequence with GELU activations and 0.1 dropout. Third, an SFT branch applies scale-and-shift modulation via two parallel 1D convolution stacks ($256 \rightarrow 256 \rightarrow 256$, kernel size 1), using the mean-pooled text condition for modulation. Fourth, an 8-head cross-attention layer treats the SFT-modulated image tokens as queries and the original image tokens as keys/values (all projections with 256 channels). During training, a contrastive alignment branch further maps pooled image and text features into a shared 256-dimensional space.

Overall, GeoMAR contains 124.51M parameters, of which 51.83M are trainable under this freezing strategy, as summarized in Table~\ref{tab:param_breakdown}.
\begin{table}[t]
\centering
\footnotesize
\setlength{\tabcolsep}{4.5pt}
\renewcommand{\arraystretch}{1.1}
\caption{GeoMAR parameter breakdown.}
\label{tab:param_breakdown}
\begin{tabular}{lccc}
\toprule
\textbf{Module} & \textbf{Structure} & \textbf{Params} & \textbf{Trainable} \\
\midrule
Frozen VQ backbone & VQGAN (HQ/LQ) + codebook & 102.16M & 29.48M \\
MAR transformer & 9$\times$SA blocks, dim 512 & 20.37M & 20.37M \\
Injector & projector + SFT + cross-attn  & 1.97M & 1.97M \\
\midrule
Total & -- & 124.51M & 51.83M \\
\bottomrule
\end{tabular}
\end{table}

\begin{table}[t]
\centering
\footnotesize
\setlength{\tabcolsep}{4.5pt}
\renewcommand{\arraystretch}{1.1}
\caption{Comparison of model complexity and reported inference time. For GeoMAR, $^\dagger$ denotes the online restoration-core latency.}

\label{tab:complexity}
\resizebox{\columnwidth}{!}{
\begin{tabular}{llccc}
\toprule
Prior type & Methods & Params (M) & FLOPs (G) & Time (s) \\
\midrule
GAN-based & GFPGAN & 60.76 & 170.00 & 0.13 \\
 & GPEN & 71.01 & 137.06 & 0.06 \\
\midrule
Diffusion-based & DR2 & 85.79 & 725.08 & 2.05 \\
 & DifFace & 175.40 & 185.95 & 7.05 \\
 & VSPBFR & 420.88 & 570.21 & 0.11 \\
 & DiffusionReward & 2045.69 & 44949.94 & 4.23 \\
\midrule
Codebook-based & VQFR & 79.30 & 1012.94 & 5.71 \\
 & CodeFormer & 84.22 & 802.54 & 0.07 \\
 & DAEFR & 141.68 & 1110.97 & 0.12 \\
 & GeoMAR (Ours) & 124.51 & 702.49 & 0.24$^\dagger$ \\
\bottomrule
\end{tabular}}
\end{table}

 \subsection{Model Complexity and Inference Efficiency}

To provide a comprehensive assessment of efficiency, we evaluate both the computational complexity and the inference runtime of our framework. As shown in Table~\ref{tab:complexity}, we report the number of parameters (Params), computational cost (FLOPs), and online restoration time. The GeoMAR core achieves a favorable balance between restoration performance and efficiency, requiring fewer parameters and lower FLOPs than the competitive baseline DAEFR. We further provide a stage-by-stage latency profile in Table~\ref{tab:e2e_runtime_supp}. For a newly observed input, BiSeNet face parsing takes $0.03$ seconds, Qwen3-VL-8B prompt generation takes $3.63$ seconds, Flan-T5 text encoding takes $0.01$ seconds, and MAR prediction plus final decoding takes $0.24$ seconds, giving a complete end-to-end latency of $3.91$ seconds per image. The first three stages generate static conditioning information and can be pre-computed or cached. Nevertheless, they are included in the end-to-end total for transparency. Accordingly, the $0.24$-second entry in Table~\ref{tab:complexity} refers specifically to the online GeoMAR restoration core after the geometric prior has been obtained. Under this inclusive definition, GeoMAR remains faster than the reported $7.05$-second runtime of DifFace while benefiting from component-level multimodal guidance.

\section{More Discussions on GeoMAR}
\label{sec1}

\subsection{CodeFormer with Masked Auto-regressive Token Prediction}

While the main paper evaluates the effectiveness of the masked auto-regressive token prediction (MAR) module within DAEFR, we further validate its plug-and-play capability by integrating it into another baseline method, CodeFormer. Quantitative results depicted in Table~\ref{compare_code} show improvements on LFW-Test datasets by applying MAR, while achieving comparable performance on the WebPhoto-Test dataset. The visual comparisons showed in Fig.~\ref{code} reveal that MAR enables better restoration through iterative refinement over the original single-step approach. Notably, even in cases where the quantitative metrics are comparable, the results with MAR exhibit visibly improved fine-grained details, particularly in challenging regions such as hair structures.

\subsection{Ablation Study of the Contrastive Alignment Loss}
To evaluate the contribution of the contrastive alignment loss ($\mathcal{L}_{contra}$) independently of the Aligned Geometric Priors Injector, we train a variant with $\mathcal{L}_{contra}$ removed while keeping all architectures and hyperparameters unchanged. As shown in Table~\ref{tab:ablation_lcontra}, removing $\mathcal{L}_{contra}$ consistently degrades FID across both datasets and results in worse or comparable NIQE. In the injector, the projected textual-prior embeddings $Z_{txt}$ modulate the original LQ features $Z_{LQ}$ through SFT. Under severe degradation, the raw LQ features can exhibit a substantial semantic disparity from high-level textual priors. The contrastive objective reduces this cross-modal gap before SFT modulation, enabling the injector to produce more accurate and geometrically faithful aligned features $Z_{\mathrm{align}}$.
\begin{table}[t]
  \centering
  \footnotesize
  \setlength{\tabcolsep}{6pt}
  \caption{Quantitative results of CodeFormer with and without MAR on two real-world datasets (best \textbf{FID} in \textcolor{red}{red}).}
  \label{compare_code}
  {
  \resizebox{\linewidth}{!}{
    \begin{tabular}{@{} llcc @{}}
      \toprule
      Model & Methods & LFW-Test & WebPhoto-Test \\
      \midrule
      \multirow{2}{*}{CodeFormer}
        & w/o MAR & 52.840 & \textcolor{red}{83.936} \\
        & w/ MAR & \textcolor{red}{51.635} & 84.416 \\
      \bottomrule
    \end{tabular}
  }
  }
\end{table}

\begin{table}[t]
\centering
\setlength{\tabcolsep}{4.5pt}
\caption{Per-image end-to-end runtime breakdown of GeoMAR.}
\label{tab:e2e_runtime_supp}
\resizebox{\columnwidth}{!}{

\begin{tabular}{@{}llcc@{}}
\toprule
\textbf{Stage} & \textbf{Module} & \textbf{Time (s)} & \textbf{Cacheable?} \\
\midrule
Face parsing & BiSeNet & 0.03 & Yes \\
Prompt generation & Qwen3-VL-8B & 3.63 & Yes \\
Text embedding & Flan-T5-Large & 0.01 & Yes \\
MAR prediction + decoding & GeoMAR core ($T=8$) & 0.24 & No \\
\midrule
\textbf{Uncached end-to-end} & \textbf{Full pipeline} & \textbf{3.91} & -- \\
\bottomrule
\end{tabular}}
\end{table}

\begin{table}[t]
\footnotesize
\centering
\caption{Ablation study on the effect of the contrastive alignment loss ($\mathcal{L}_{contra}$).}
\label{tab:ablation_lcontra}
\resizebox{\linewidth}{!}{
\begin{tabular}{l c c c c}
\toprule
\multirow{2}{*}{Variant} & \multicolumn{2}{c}{LFW-Test} & \multicolumn{2}{c}{WebPhoto-Test} \\
\cmidrule(lr){2-3} \cmidrule(lr){4-5}
& FID $\downarrow$ & NIQE $\downarrow$ & FID $\downarrow$ & NIQE $\downarrow$ \\
\midrule
GeoMAR w/o $\mathcal{L}_{contra}$ &47.68  & \textcolor{red}{3.543} &77.05  & 3.971 \\
GeoMAR (Full) & \textcolor{red}{46.87} & 3.547 & \textcolor{red}{75.73} & \textcolor{red}{3.930} \\
\bottomrule
\end{tabular}
}
\end{table}

\subsection{Generalization under Extreme Real-World Degradations}
The main paper evaluates GeoMAR on three commonly used real-world BFR benchmarks. To further assess generalization under more extreme conditions, we evaluate it on SCface Distance-3, a surveillance setting characterized by severe low resolution, blur, compression artifacts, and illumination variation. As reported in Table~\ref{tab:scface_supp}, GeoMAR achieves the best FID and MANIQA among the compared methods, while DAEFR obtains the best NIQE. The results indicate improved perceptual realism and human-aligned quality under challenging surveillance degradation, although GeoMAR is not uniformly superior on every no-reference metric.

\subsection{Robustness to Imperfect Geometric Priors}
GeoMAR does not assume perfect parsing maps or VLM descriptions. The extracted geometric priors act as soft, high-level cues for disambiguating degraded visual features rather than hard reconstruction constraints. In particular, the KV--Q exchange uses prior-aligned features as queries while retaining the original LQ features as keys and values. The injector therefore retrieves evidence from the observed LQ image instead of blindly following a potentially inaccurate prior.

We quantify this robustness through a controlled study with four settings: GT parsing, predicted parsing used by GeoMAR, component-dropout parsing, and no geometric prior. Component dropout removes key facial components, including the eyes, mouth, and facial contour, from the parsing map before prompt generation, thereby simulating parsing failures and inaccurate component descriptions. As shown in Table~\ref{tab:prior_robustness_supp}, predicted parsing remains close to the GT-parsing upper bound on CelebA-Test. Although component dropout degrades performance, it retains better FID and competitive no-reference quality relative to removing the prior entirely on CelebA-Test, and outperforms the no-prior variant across all three metrics on the more challenging WIDER-Test. These results show that inaccurate priors reduce the benefit of geometric guidance but do not cause cascading restoration failure.

\subsection{Fine-Grained Ablation of Core Components}
The main paper evaluates the component-level prompt, KV--Q exchanged injector, and MAR refinement separately. To further isolate their contributions, we construct the additive ablation ladder in Table~\ref{tab:fine_grained_ablation_supp}. SFT and the projected T5 features are retained as the common prior-encoding backbone. Starting from M0, we progressively add dual-input prior extraction, component-based descriptions, KV--Q exchange, and MAR refinement. FID improves from 43.44 in M0 to 39.61 in M4, while NIQE and MANIQA remain competitive, validating the complementary effects of geometric-prior extraction, feature alignment, and iterative code refinement.

\subsection{Sensitivity to the Choice of VLM}
We replace Qwen3-VL with InternVL while keeping the remaining pipeline unchanged. Table~\ref{tab:vlm_sensitivity_supp} shows only marginal variations across all three real-world benchmarks. This limited sensitivity indicates that the performance gains primarily arise from the component-based geometric-prior extraction and alignment mechanism rather than a particular VLM backbone.

 \begin{figure}
     \centering 
 
     \includegraphics[width=\linewidth]{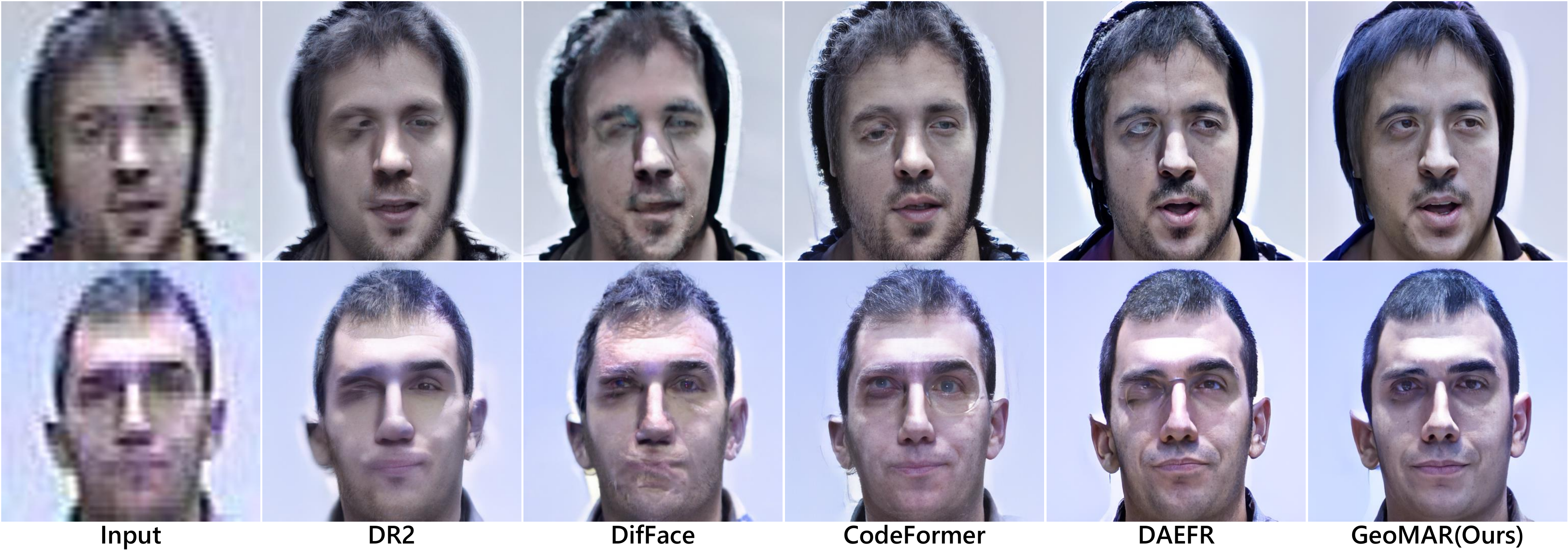}
\captionsetup{skip=2pt}
\captionof{figure}{Extreme cases on SCface Distance-3.}\label{fig:extreme_case}
     \label{fig:placeholder}
     
 \end{figure}

\begin{table}[t]
\centering
\footnotesize
\setlength{\tabcolsep}{4pt}
\caption{Quantitative comparison on SCface Distance-3.}
\label{tab:scface_supp}
\resizebox{\columnwidth}{!}{%
\begin{tabular}{lccccc}
\toprule
Metric & DR2 & DifFace & CodeFormer & DAEFR & \textbf{GeoMAR} \\
\midrule
FID $\downarrow$ & 79.69 & 91.11 & 81.87 & 82.00 & \textbf{78.80} \\
NIQE $\downarrow$ & 4.400 & 4.490 & 4.301 & \textbf{3.529} & 3.744 \\
MANIQA $\uparrow$ & 0.415 & 0.263 & 0.431 & 0.451 & \textbf{0.471} \\
\bottomrule
\end{tabular}}
\end{table}

\begin{table}[t]
\centering
\footnotesize
\setlength{\tabcolsep}{3.5pt}
\caption{Robustness to imperfect geometric priors. ``Dropout'' removes key facial components from the parsing map before prompt generation.}
\label{tab:prior_robustness_supp}
\resizebox{\columnwidth}{!}{%
\begin{tabular}{llcccc}
\toprule
Dataset & Metric & GT parsing & Predicted parsing & Dropout & No prior \\
\midrule
\multirow{3}{*}{CelebA-Test}
& FID $\downarrow$ & 39.07 & 39.61 & 40.86 & 42.36 \\
& NIQE $\downarrow$ & 3.779 & 3.819 & 3.842 & 3.829 \\
& MANIQA $\uparrow$ & 0.541 & 0.540 & 0.540 & 0.539 \\
\midrule
\multirow{3}{*}{WIDER-Test}
& FID $\downarrow$ & -- & 37.37 & 39.25 & 40.58 \\
& NIQE $\downarrow$ & -- & 3.762 & 3.813 & 3.858 \\
& MANIQA $\uparrow$ & -- & 0.510 & 0.509 & 0.500 \\
\bottomrule
\end{tabular}}
\end{table}
\begin{table}[t]
\centering
\footnotesize
\setlength{\tabcolsep}{3pt}
\caption{Fine-grained additive ablation on CelebA-Test. Dual, Comp., and KV--Q denote dual-input prior extraction, component-based descriptions, and KV--Q exchange, respectively.}
\label{tab:fine_grained_ablation_supp}
\resizebox{\columnwidth}{!}{%
\begin{tabular}{lccccccc}
\toprule
Setting & Dual & Comp. & KV--Q & MAR & FID $\downarrow$ & NIQE $\downarrow$ & MANIQA $\uparrow$ \\
\midrule
M0 & \no & \no & \no & \no & 43.44 & 3.962 & 0.534 \\
M1 & \yes & \no & \no & \no & 42.64 & 3.927 & 0.539 \\
M2 & \yes & \yes & \no & \no & 42.59 & 3.870 & 0.542 \\
M3 & \yes & \yes & \yes & \no & 42.23 & 3.891 & 0.540 \\
\textbf{M4 (GeoMAR)} & \yes & \yes & \yes & \yes & \textbf{39.61} & 3.819 & 0.540 \\
\bottomrule
\end{tabular}}
\end{table}


\begin{table}[t]
\centering
\setlength{\tabcolsep}{4pt}
\caption{Sensitivity to the VLM used for geometric-prior extraction.}
\label{tab:vlm_sensitivity_supp}
\begin{tabular}{llcc}
\toprule
Dataset & Metric & Qwen3-VL & InternVL \\
\midrule
\multirow{3}{*}{LFW-Test}
& FID $\downarrow$ & 46.87 & 46.44 \\
& NIQE $\downarrow$ & 3.547 & 3.583 \\
& MANIQA $\uparrow$ & 0.540 & 0.539 \\
\midrule
\multirow{3}{*}{WebPhoto-Test}
& FID $\downarrow$ & 75.73 & 75.72 \\
& NIQE $\downarrow$ & 3.930 & 3.940 \\
& MANIQA $\uparrow$ & 0.499 & 0.499 \\
\midrule
\multirow{3}{*}{WIDER-Test}
& FID $\downarrow$ & 37.37 & 37.63 \\
& NIQE $\downarrow$ & 3.762 & 3.766 \\
& MANIQA $\uparrow$ & 0.510 & 0.511 \\
\bottomrule
\end{tabular}
\end{table}

\section{More Visual Results}
\label{sec3}

In this section, we provide more qualitative results to further demonstrate the superiority of our proposed GeoMAR and validate the effectiveness of its core components.

\subsection{Comparisons on different textual priors} 
In Fig.~\ref{geo}, we further compare textual prior extraction strategies, including w/o Geo-Priors, concat-style priors, and our component-based geometric priors (Ours). The results indicate that our dual-input explicit priors significantly reduce semantic hallucinations and provide more faithful structural guidance.

\subsection{Effectiveness of KV-Q Exchange Injector} 
In Fig.~\ref{qkv}, we compare different query designs in the cross-attention module, including raw and enhanced LQ features as queries. Compared with raw LQ queries, our KV-Q exchange design selectively extracts high-fidelity information, enabling better recovery of fine facial details while preserving input-consistent attributes.

\subsection{Comparisons on different prediction steps $T$ in MAR} 
In Fig.~\ref{step}, we provide qualitative results generated using different auto-regressive prediction steps ($T = 1, 4, 8, 12$) during inference. The results at $T$ = 8 exhibit richer texture details and higher fidelity. Although the inference time of $T$ = 8 is slightly higher than that of $T$ = 1, it achieves the best overall restoration quality, making it a favorable trade-off between performance and efficiency.
\subsection{Effectiveness of the MAR module} 

To demonstrate the plug-and-play capability of MAR token prediction, we provide visual comparisons with and without MAR. Fig.~\ref{code} and Fig.~\ref{daefr} show the improvements when MAR is integrated into CodeFormer and DAEFR, while Fig.~\ref{mar} compares our GeoMAR with and without MAR. The results demonstrate that multi-step MAR refinement effectively reduces structural distortions and artifacts from fragile one-step mapping.

\subsection{Comparisons with representative methods} 
We further provide visual comparisons between GeoMAR and representative blind face restoration methods on multiple datasets. Fig.~\ref{supp_lfw}, Fig.~\ref{supp_web}, Fig.~\ref{supp_wider}, and Fig.~\ref{supp_cele} present results on LFW-Test, WebPhoto-Test, WIDER-Test, and CelebA-Test, respectively. Under diverse real-world and synthetic degradations, GeoMAR consistently restores more realistic details and coherent structures than competing methods.
\section{Limitations}
\paragraph{Failure cases.}
\begin{figure}
    \centering
    \includegraphics[width=\linewidth]{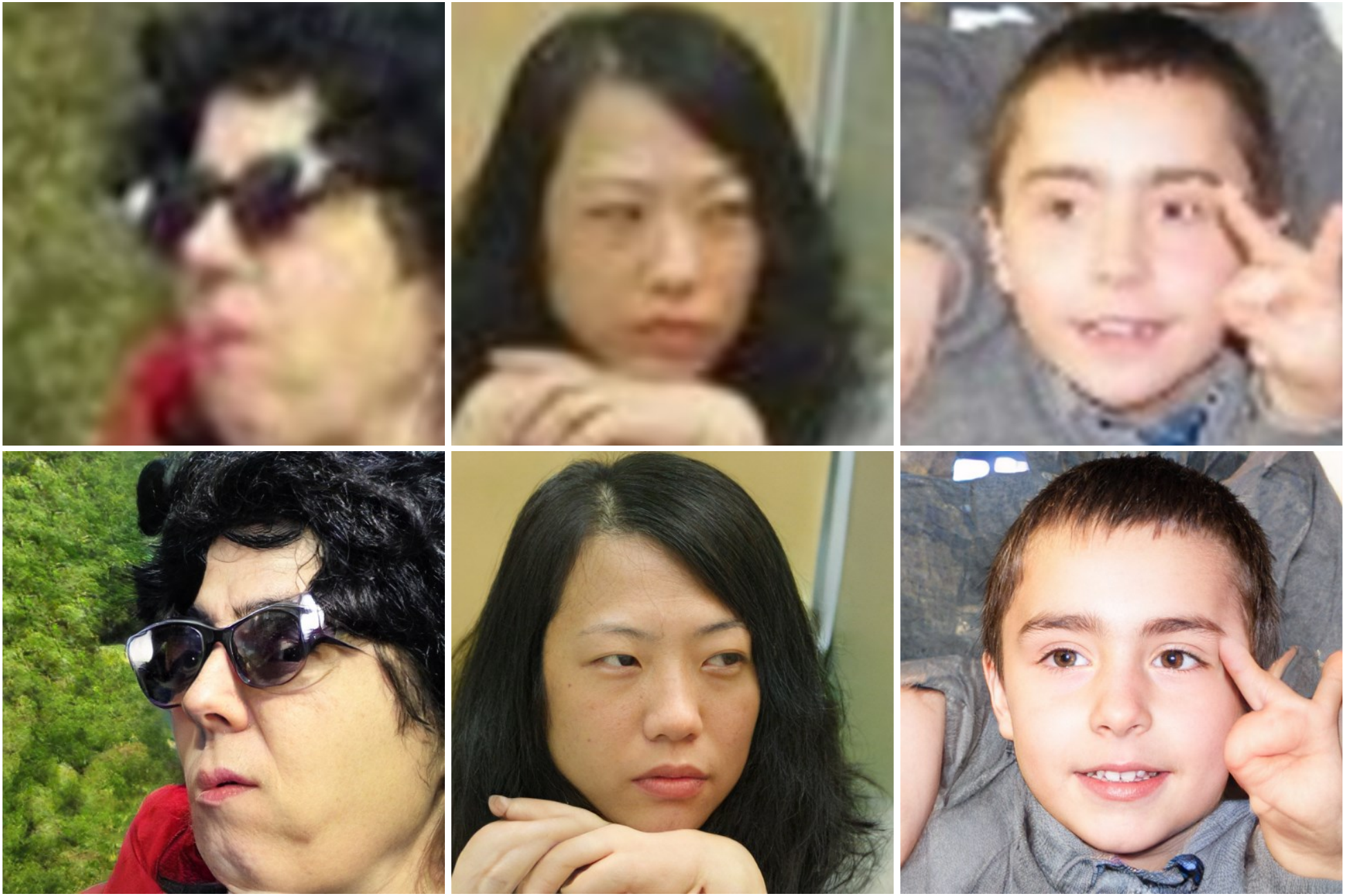}
       \caption{Failure cases under severe occlusions. }
    \label{fig:failure_cases}
\end{figure}
Figure~\ref{fig:failure_cases} shows failures caused by sunglasses, hands, and foreground objects. These occlusions corrupt both RGB evidence and parsing maps, leading to inaccurate descriptions and input-inconsistent restorations. When identity-related evidence is fully hidden, the restored face may alter occluded attributes or remove parts of the occluder.

Although GeoMAR demonstrates strong performance, our model is built upon a pre-trained codebook, and its reconstruction quality is inherently limited by the capacity of codebook and the diversity of HQ features it encodes. As a result, it struggles to accurately reconstruct fine-grained accessories such as earrings and hair ornaments and complex background.


\begin{figure*}[h]
  \centering
  \begin{minipage}[t]{1\textwidth}
    \centering
    \includegraphics[width=\linewidth]{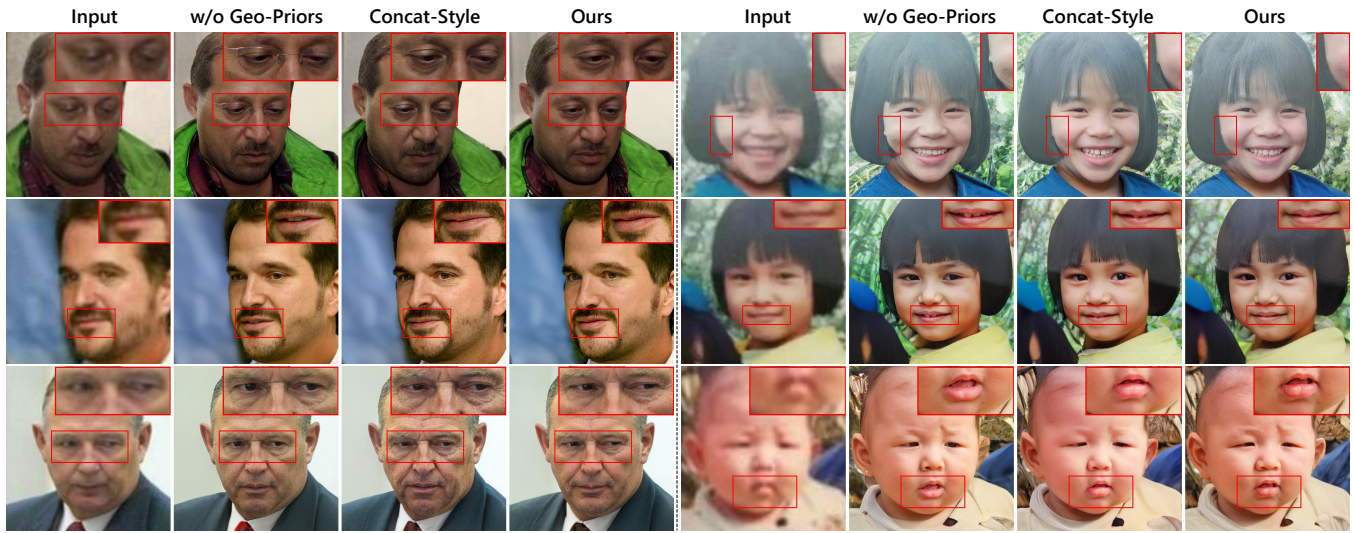}
    \captionof{figure}{Qualitative comparison of different textual priors on two real-world datasets.}
    \label{geo}
  \end{minipage}
\end{figure*}

\begin{figure*}[htbp] 
    \centering 
    \includegraphics[width=1\textwidth]{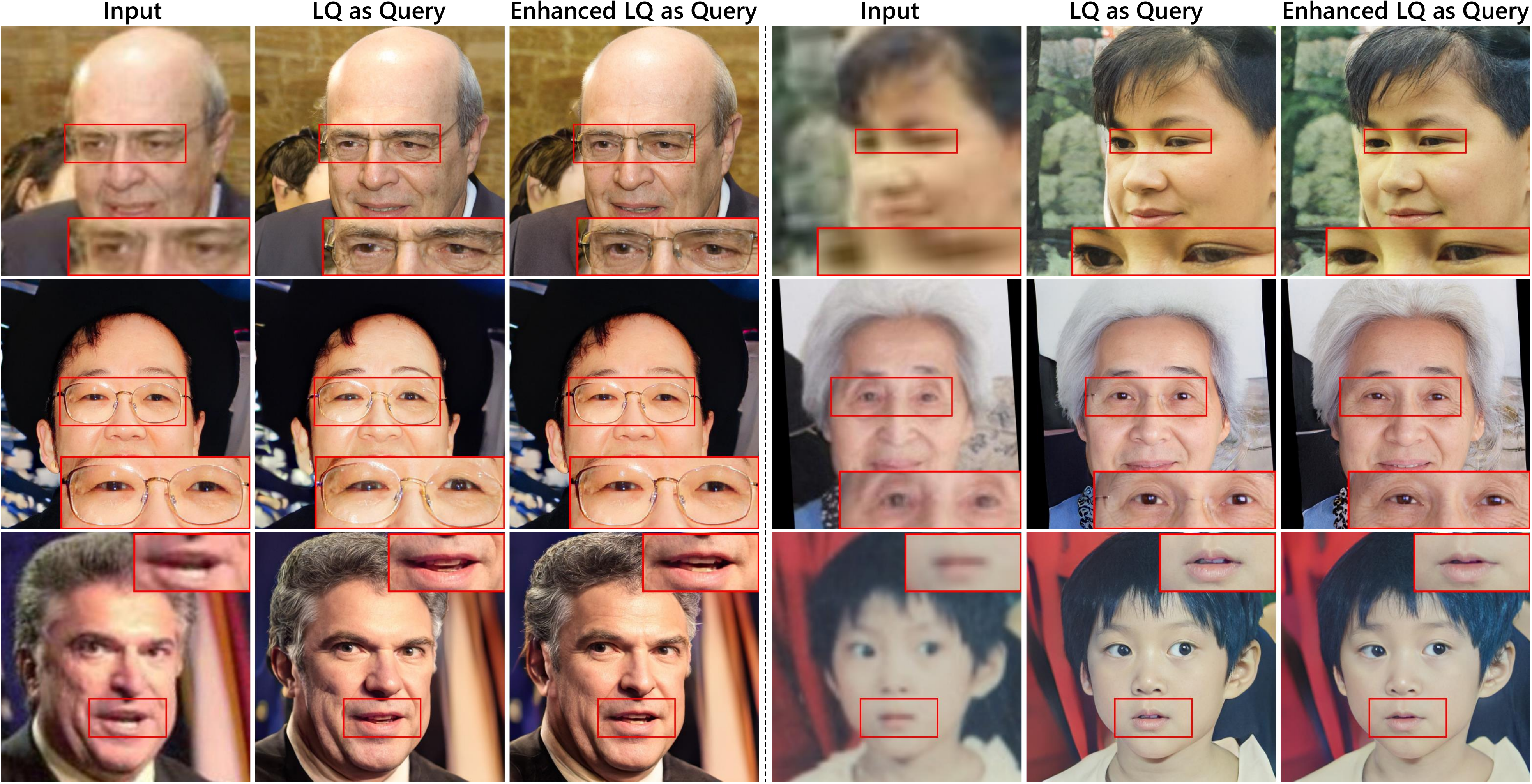} 
    \caption{Qualitative comparisons of different types of query on two real-world datasets.} 
    \label{qkv} 
\end{figure*}

\begin{figure*}[h]
  \centering
  \begin{minipage}[t]{1\textwidth}
    \centering
    \includegraphics[width=\linewidth]{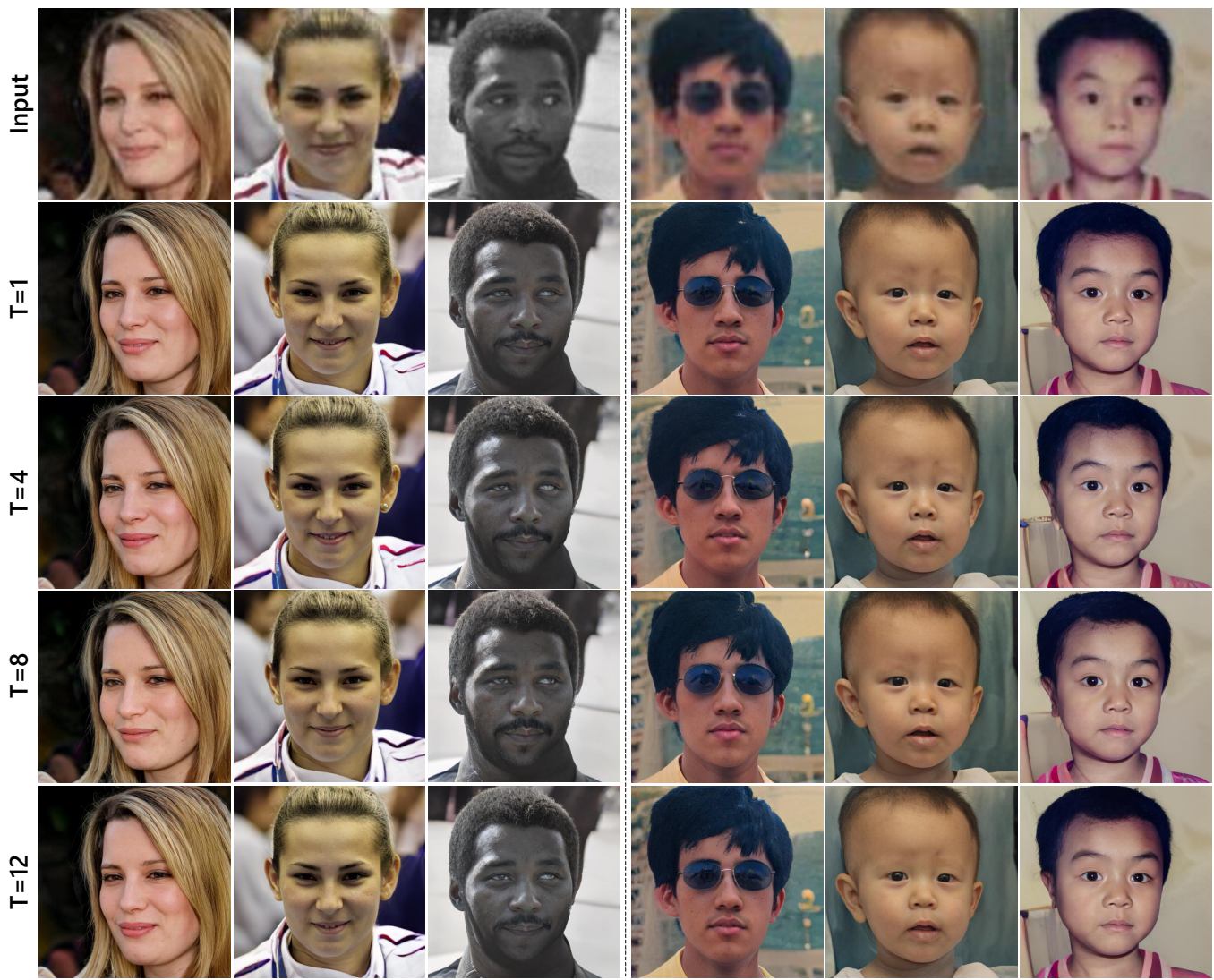}
    \captionof{figure}{Qualitative comparisons of different steps $T$ in masked auto-regressive on two real-world datasets.}
    \label{step}
  \end{minipage}
\end{figure*}

\begin{figure*}[h]
  \centering
  \begin{minipage}[t]{1\textwidth}
    \centering
    \includegraphics[width=\linewidth]{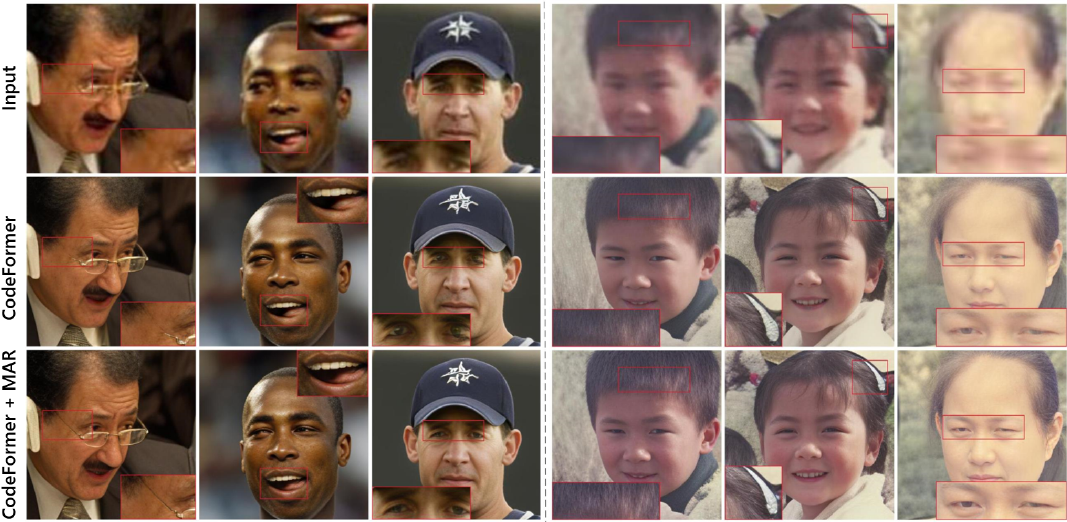}
    \captionof{figure}{Qualitative comparisons of CodeFormer with and without MAR on two real-world datasets.}
    \label{code}
  \end{minipage}
\end{figure*}

\begin{figure*}[h]
  \centering
  \begin{minipage}[t]{1\textwidth}
    \centering
    \includegraphics[width=\linewidth]{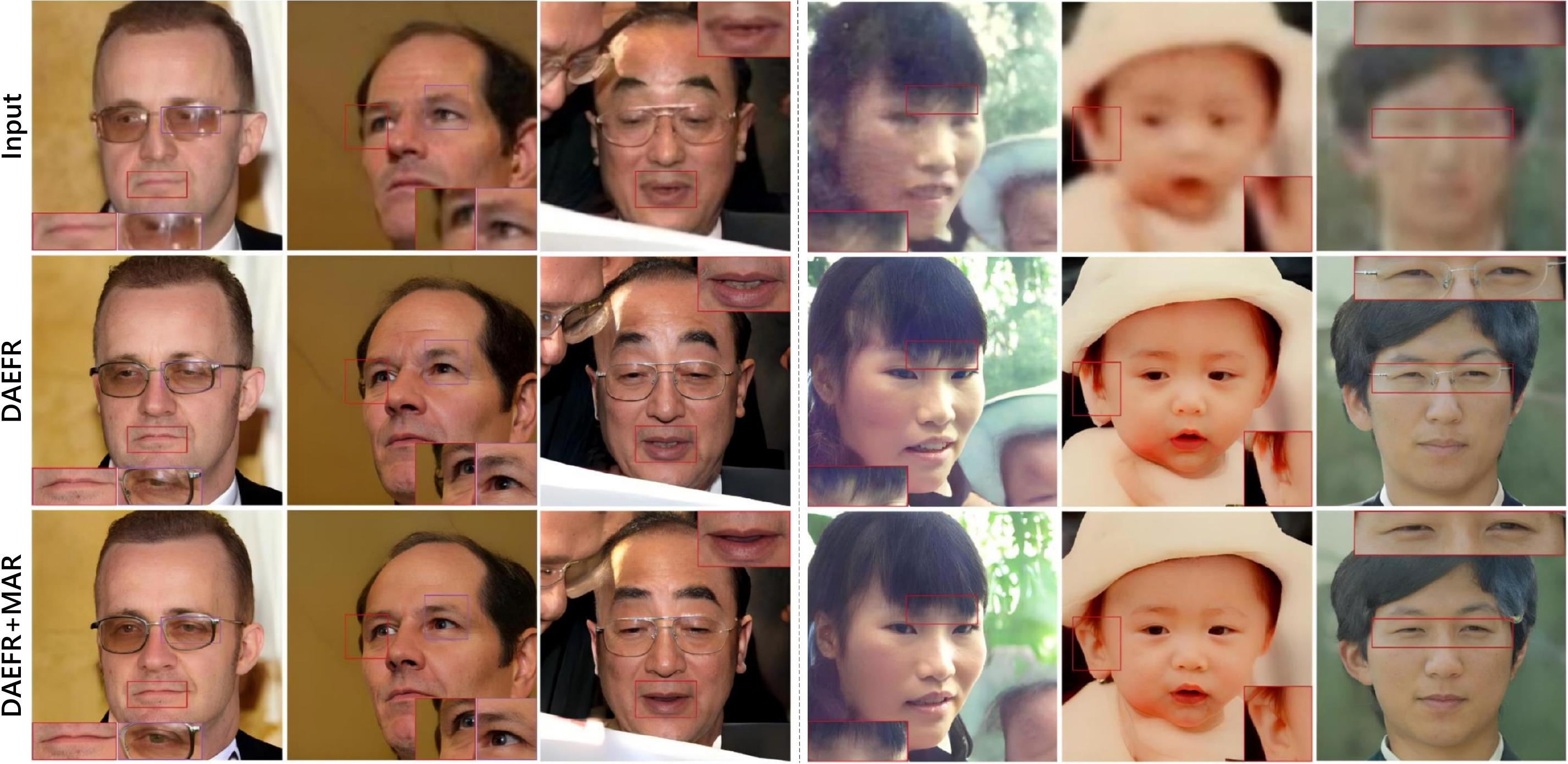}
    \captionof{figure}{Qualitative comparisons of DAEFR with and without MAR on two real-world datasets.}
    \label{daefr}
  \end{minipage}
\end{figure*}

\begin{figure*}[h]
  \centering
  \begin{minipage}[t]{1\textwidth}
    \centering
    \includegraphics[width=\linewidth]{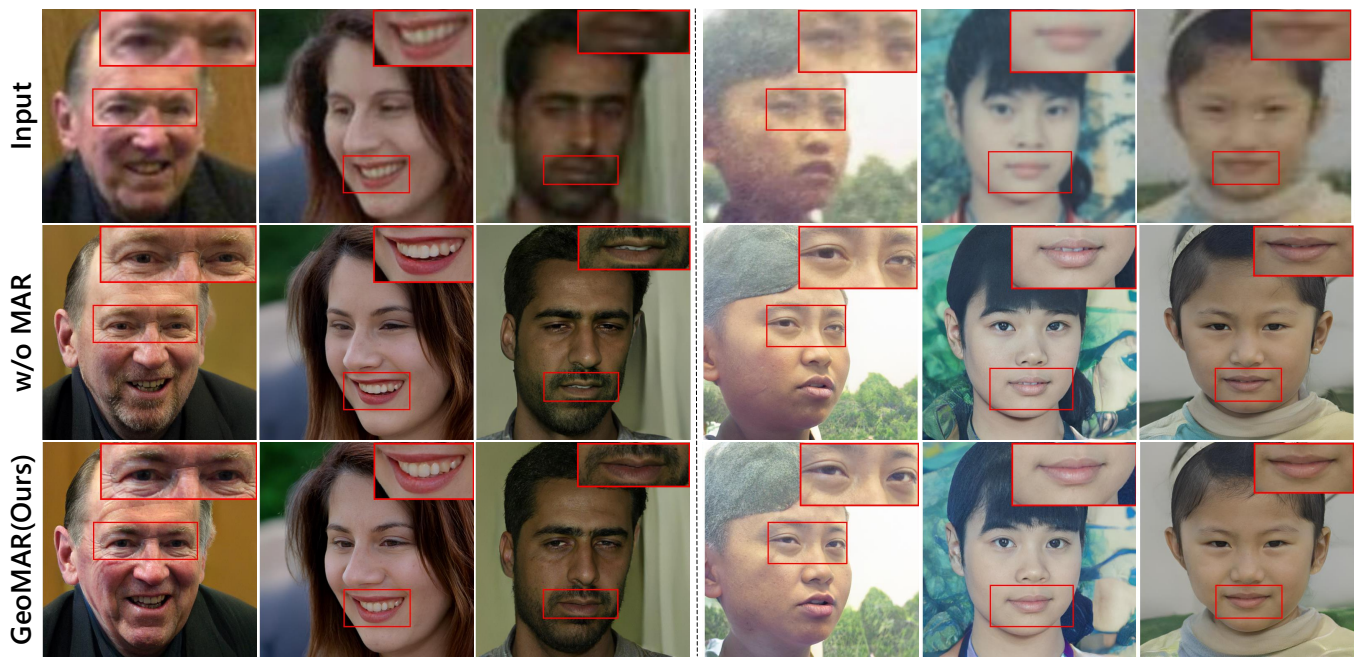}
    \captionof{figure}{Qualitative comparisons of GeoMAR with and without MAR on two real-world datasets.}
    \label{mar}
  \end{minipage}
\end{figure*}

\begin{figure*}[htbp] 
    
  \centering 
    \includegraphics[width=1\textwidth]{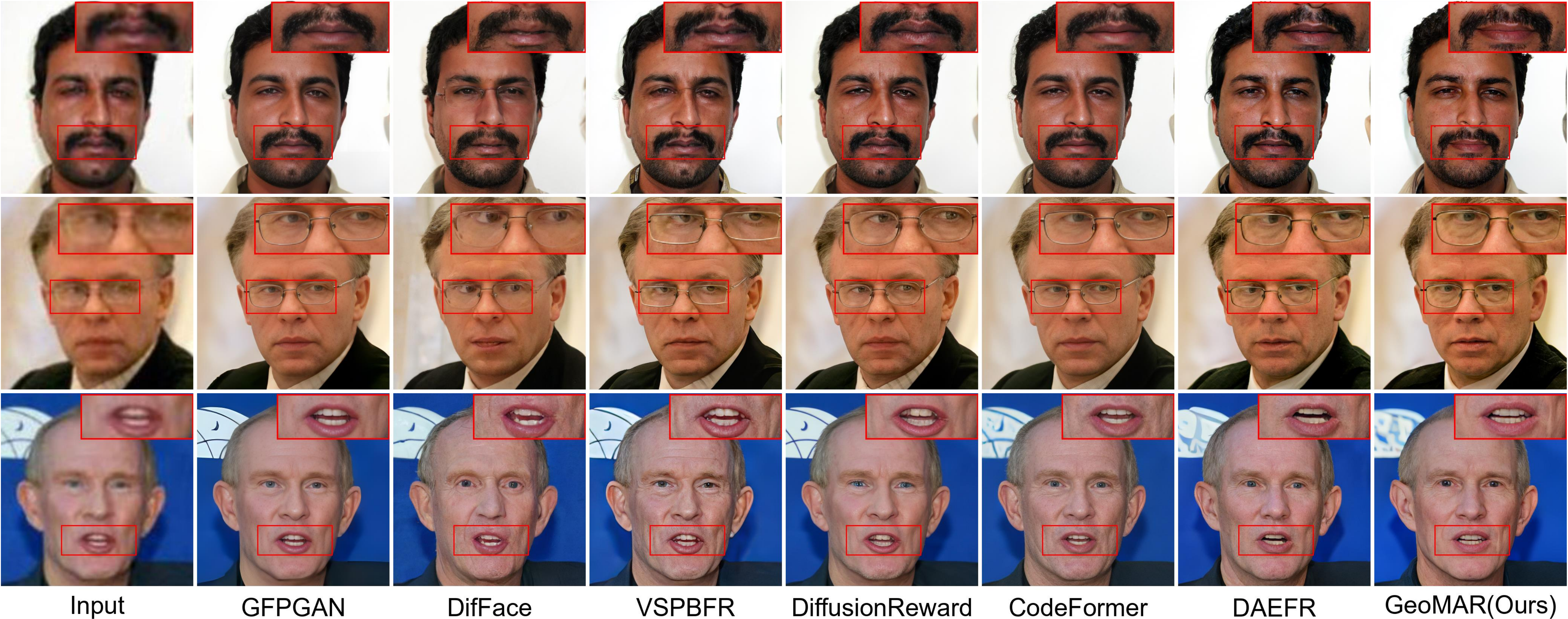} 
     \caption{Visual comparisons of different methods on LFW-Test dataset.} 
    \label{supp_lfw}
\end{figure*}

\begin{figure*}[htbp] 
    \centering 
    \includegraphics[width=1\textwidth]{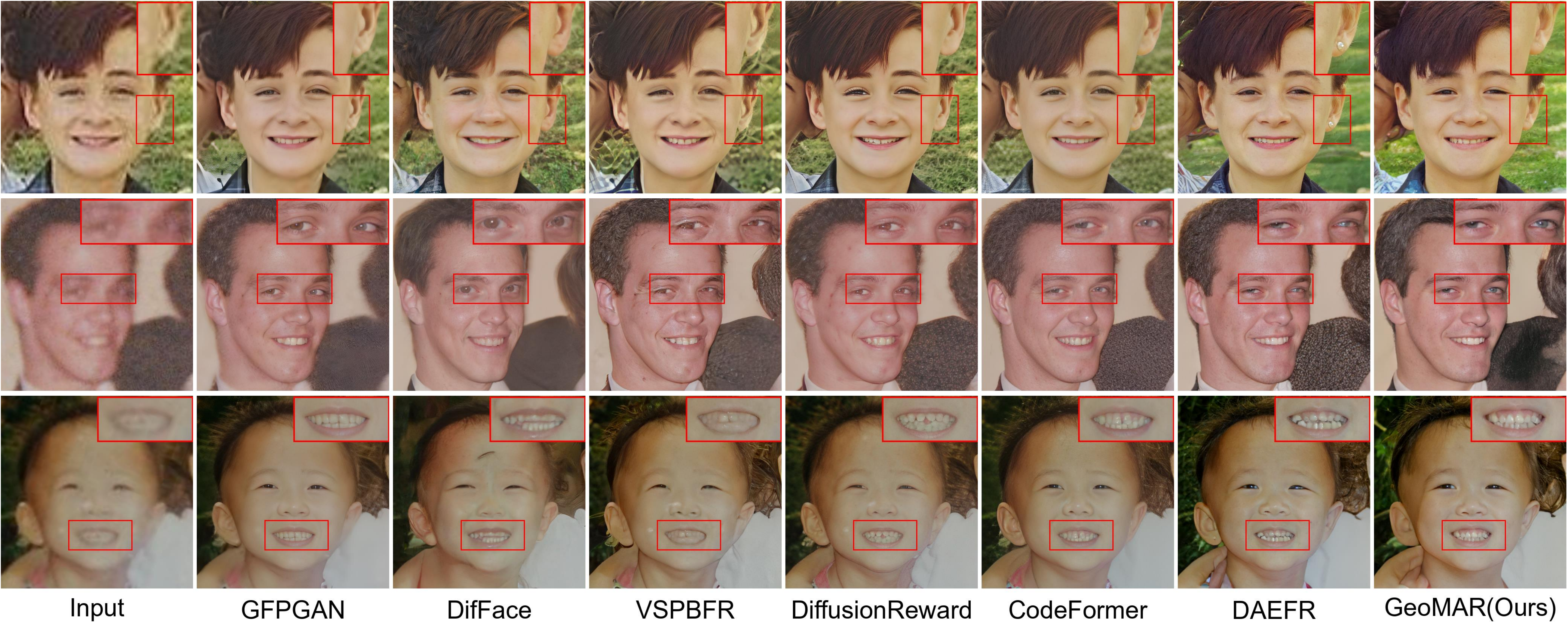} 
        \caption{Visual comparisons of different methods on WebPhoto-Test dataset.} 
        \label{supp_web}
\end{figure*}

\begin{figure*}[htbp] 
    \centering 
    \includegraphics[width=1\textwidth]{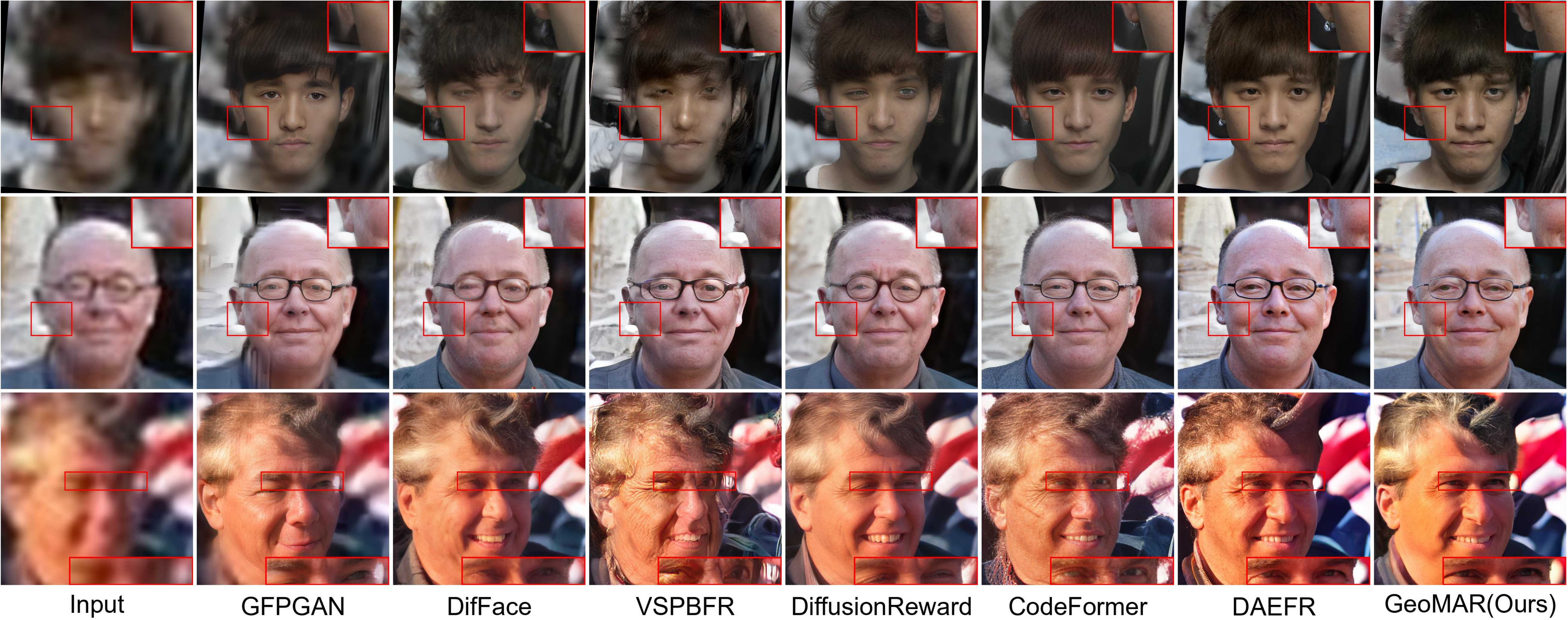} 
        \caption{Visual comparisons of different methods on WIDER-Test dataset.} 
        \label{supp_wider}
\end{figure*}

\begin{figure*}[htbp] 
    \centering 
    \includegraphics[width=1\textwidth]{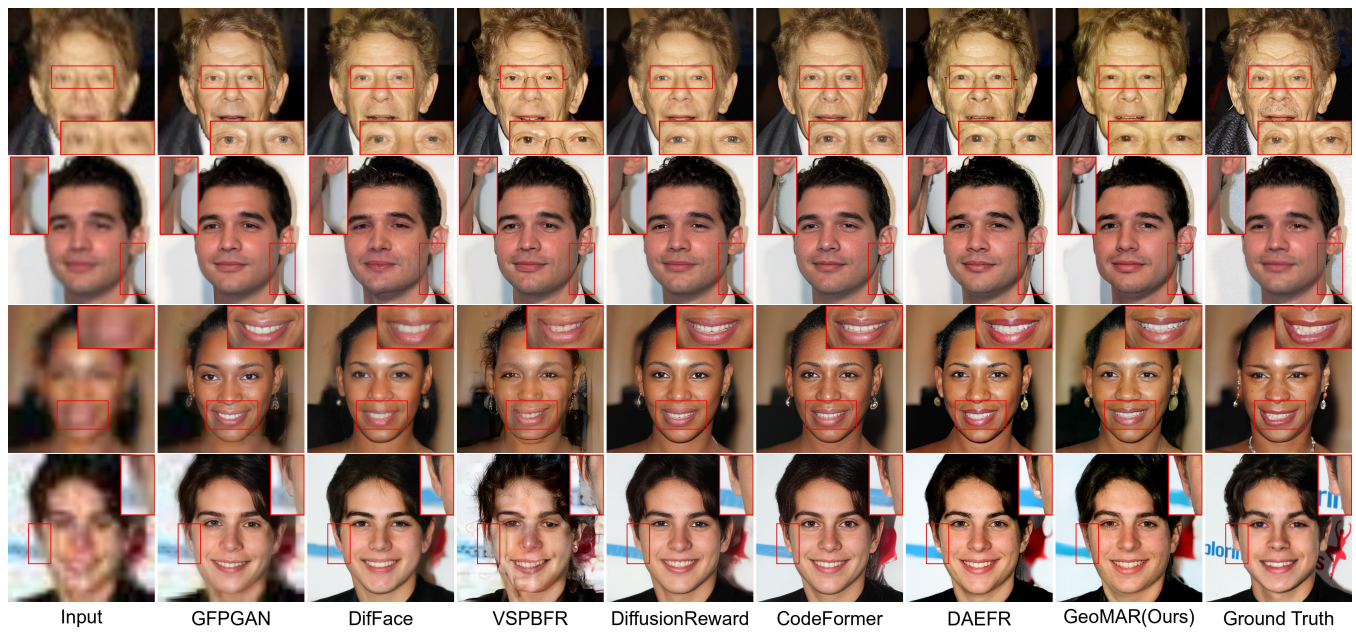} 
        \caption{Visual comparisons of different methods on synthetic CelebA-Test dataset.} 
        \label{supp_cele}
\end{figure*}



\end{document}